\documentclass[journal]{IEEEtran}
\usepackage{soul}
\usepackage{color}
\usepackage{amsmath}
\usepackage{graphicx}
\usepackage{bm}
\usepackage{amssymb}
\usepackage{physics}
\usepackage{algorithm}
\usepackage{algpseudocode}

\usepackage{url}
\usepackage{verbatim}
\usepackage{rotating}
\usepackage{array}
\usepackage{booktabs}
\usepackage{multirow}
\usepackage{makecell}
\usepackage{underscore}
\usepackage[english]{babel}
\usepackage{xcolor,colortbl}
\usepackage[numbers,sort&compress]{natbib}

\DeclareMathOperator{\E}{\mathbb{E}}
\DeclareMathOperator*{\argmin}{arg\,min}

\usepackage{xr}
\begin{document}
%

\title{Evolutionary Optimization of Physics-Informed Neural Networks: Advancing Generalizability by the Baldwin Effect}
%
%
%


\author{Jian Cheng Wong, Chin Chun Ooi, Abhishek Gupta, \textit{Senior Member, IEEE}, Pao-Hsiung Chiu, \\ Joshua Shao Zheng Low, My Ha Dao, and Yew-Soon Ong, \textit{Fellow, IEEE}
\thanks{Jian Cheng Wong, Chin Chun Ooi, and Pao-Hsiung Chiu are with the Institute of High Performance Computing, Agency for Science, Technology and Research, Singapore (e-mail: wongj@a-star.edu.sg; ooicc@a-star.edu.sg; chiuph@a-star.edu.sg).}
\thanks{Abhishek Gupta is with the School of Mechanical Sciences, Indian Institute of Technology Goa, India (e-mail: abhishekgupta@iitgoa.ac.in).}
\thanks{Joshua Shao Zheng Low is with the College of Computing and Data Science, Nanyang Technological University, Singapore (e-mail: joshualow188@gmail.com).}
\thanks{My Ha Dao is with the Technology Centre for Offshore and Marine, Singapore (e-mail: Dao_My_Ha@tcoms.sg).}
\thanks{Yew-Soon Ong is with the Agency for Science, Technology and Research, Singapore, and is also with the College of Computing and Data Science, Nanyang Technological University, Singapore (e-mail: Ong_Yew_Soon@a-star.edu.sg).}}

%
%

\maketitle

\begin{abstract}
Physics-informed neural networks (PINNs) are at the forefront of scientific machine learning, making possible the creation of machine intelligence that is cognizant of physical laws and able to accurately simulate them. However, today's PINNs are often trained for a single physics task and require computationally expensive re-training for each new task, even for tasks from similar physics domains. To address this limitation, this paper proposes a pioneering approach to advance the generalizability of PINNs through the framework of Baldwinian evolution. Drawing inspiration from the neurodevelopment of precocial species that have evolved to learn, predict and react quickly to their environment, we envision PINNs that are pre-wired with connection strengths inducing strong biases towards efficient learning of physics. A novel two-stage stochastic programming formulation coupling evolutionary selection pressure (based on proficiency over a distribution of physics tasks) with lifetime learning (to specialize on a sampled subset of those tasks) is proposed to instantiate the Baldwin effect. The evolved Baldwinian-PINNs demonstrate fast and physics-compliant prediction capabilities across a range of empirically challenging problem instances with more than an order of magnitude improvement in prediction accuracy at a fraction of the computation cost compared to state-of-the-art gradient-based meta-learning methods. For example, when solving the diffusion-reaction equation, a 70x improvement in accuracy was obtained while taking 700x less computational time. This paper thus marks a leap forward in the evolutionary meta-learning of PINNs as generalizable physics solvers. Sample codes are available at \url{https://github.com/chiuph/Baldwinian-PINN}.

\end{abstract}

\begin{IEEEkeywords}
Baldwin effect, evolutionary optimization, neuroevolution, meta-learning, physics-informed neural networks
\end{IEEEkeywords}

%
\IEEEpeerreviewmaketitle

\section{Introduction}
%
%
%
%
\IEEEPARstart{T}{he} emerging field of scientific machine learning seeks to create more accurate, data-efficient, and explainable machine intelligence for science and engineering. The direct incorporation of mathematically expressible laws of nature into learned models to ensure physically consistent predictions is an appealing proposition, as evidenced by the proliferation of physics-informed neural networks (PINNs) across multiple scientific domains since seminal work by \textit{Raissi et al.} \cite{raissi2019physics}. The key concept is to utilize physics-based mathematical relations or constraints as a regularization loss (\textit{aka} physics-informed loss). This physics-informed loss is amenable to various forms of scientific knowledge and theories, including fundamental ordinary or partial differential equations (ODEs or PDEs). It flexibly incorporates scientific discoveries accumulated across centuries into the machine intelligence models of today across diverse scientific and engineering disciplines \cite{cuomo2022scientific, karniadakis2021physics, raissi2020hidden, huang2022applications, de2021assessing}.

However, PINNs remain limited in their ability to generalize across physics scenarios. Contrary to its promise, a PINN does not guarantee compliance with physics when used for new scenarios unseen during training, e.g., variations in PDE parameters, initial conditions (ICs) or boundary conditions (BCs) that lie outside the confines of their training. Instead, these predictions remain physics-agnostic and may experience similar negative implications for reliability as typical data-driven models.

In principle, physics-compliant predictions for any new scenario can be achieved by performing physics-informed retraining\textemdash an attractive feature of PINNs\textemdash even without labelled data. However, the additional training can be cost-prohibitive as physics-informed learning is much more difficult than data-driven learning due to the ruggedness of the loss landscapes~\cite{wang2023expert, wong2022learning,chiu2022can,wong2023lsa}. This has motivated the exploration of transfer learning techniques where connection strengths from similar (source) physics scenarios are used to facilitate accurate learning of solutions for new, harder problems \cite{wong2025evolutionaryoptimizationphysicsinformedneural,wong2021can,krishnapriyan2021characterizing}. The related notion of meta-learning seeks to discover an optimized initialization of a model to enable rapid adaptation to a new test task with minimal training \cite{wong2025evolutionaryoptimizationphysicsinformedneural}. Nonetheless, most transfer- and meta-learned PINNs proposed to date still require the aforementioned physics-informed retraining (with a substantial number of optimization iterations) to achieve more accurate solutions, and are therefore not ideal for applications that call for repeated, fast evaluations. A method to arrive at a generalizable PINN, one that can provide fast and accurate physics prediction/simulation across a varied set of unseen scenarios, remains elusive.


\begin{figure*}[h]%
\centering
\includegraphics[width=1.0\textwidth]{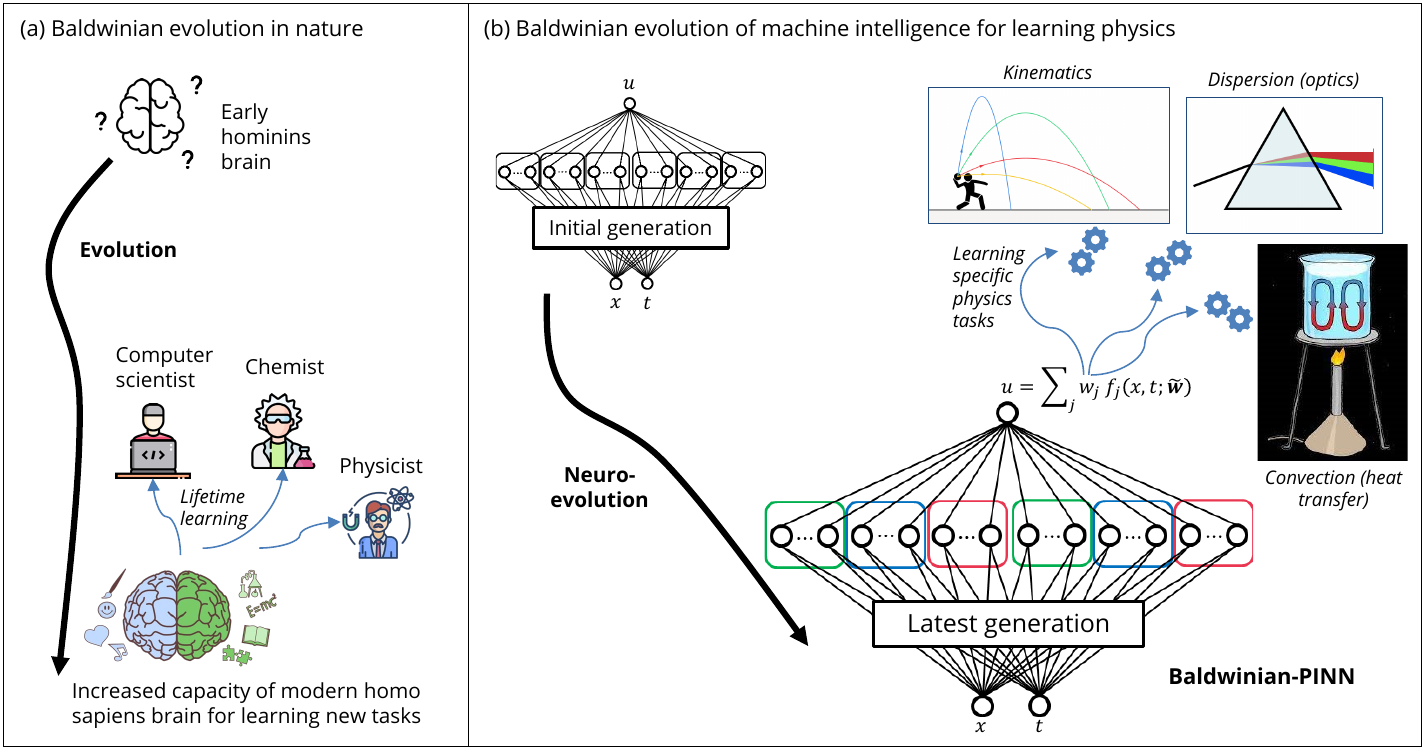}
\vspace{-0.6cm}
\caption{Schematic diagram of (a) Baldwinian evolution in nature and (b) evolving machine intelligence for learning physics with Baldwinian-PINNs. In nature, the Baldwin effect describes how learned traits are eventually reinforced in the genetic makeup of a population of organisms through natural selection. Equivalently, a population of Baldwinian-PINNs evolves over generations by being exposed to a broad distribution of physics tasks, gradually reinforcing traits promoting accurate physics learning into their genetic makeup. The evolved Baldwinian-PINNs are inherently equipped with strong learning biases to accurately solve any physics tasks over a broad task distribution.}\label{fig:schematic}
\end{figure*}


In search of machine intelligence that generalizes for physically-consistent simulations of varied processes in the natural world, this paper studies the meta-learning of PINNs through the lens of neural Baldwinism---an expression of the \emph{Baldwin effect} in the evolution of brains and intelligence~\cite{downing2012heterochronous}. Inspiration is drawn from the transmission of knowledge and predispositions across generations in precocial species, whereby their young are ``born ready" with strong learning biases to perform a wide range of tasks. In order to pre-wire such learning ability into the initial connection strengths of a neural network, we examine an algorithmic realization of neural Baldwinism in the context of PINNs. The essence of Baldwinism lies in the phenomenon that characters learned by individual organisms of a group during their lifetime may eventually, under selection pressure, get reinforced by associated hereditary characters \cite{simpson1953baldwin}. Analogously, Baldwinian neuroevolution of PINNs consists of an outer evolutionary optimization loop in which populations of PINN models are collectively exposed to a wide range of physics tasks sampled from a probability distribution over tasks of interest. Models with higher propensity to perform a random subset of those tasks well in an inner lifetime learning loop are evaluated as being fitter for survival, thus inducing a selection pressure towards connection strengths that encode stronger learning biases. The parallels between neural Baldwinism in nature and that of PINNs is depicted in Figure~\ref{fig:schematic}. The mathematical formulation of the problem resembles a \emph{two-stage stochastic program} \cite{powell2014clearing, bakker2020structuring}, where the solution to lifetime learning enables the PINN to rapidly specialize to a specific physics task at test time. 

Harnessing evolutionary procedures to optimize neural networks lends much greater versatility relative to other meta-learning approaches in terms of jointly crafting network architectures, initial network connection strengths, as well as learning algorithms, all through the use of potentially non-differentiable fitness functions \cite{fernando2018meta}. Such biological ``neuroevolution"~\cite{stanley2019designing, miikkulainen2021biological} precludes the need for explicit parameterization of tasks, facilitating generalization over task distributions comprising any broad mix of ODE/PDEs, ICs, and BCs. The physics-informed lifetime learning of the neural network can be accelerated by reduction to a least-squares learning problem in its final layer, making Baldwinian neuroevolution computationally feasible. Such a least-squares formulation yields a closed-form result by means of the Tikhonov regularization~\cite{golub1999tikhonov} and guarantees zero physics-informed loss given a sufficiently overparameterized network. The closed-form expression vastly reduces (or even eliminates) the need for iterative parameter updates to enable extremely fast lifetime learning of desired physics. Critically, the evolutionary search procedure is inherently highly parallelizable, thereby allowing for efficient meta-learning at scale by capitalizing on state-of-the-art advances in multi-CPU/GPU hardware infrastructure. 

Neural Baldwinism thus makes it possible to achieve generalizable PINNs that are ``genetically equipped" to perform well over a wide range of physics tasks. The evolved models (referred to as Baldwinian-PINNs) are demonstrated in this study to be broadly applicable to the simulation of families of linear and nonlinear ODE/PDEs encompassing diverse physical phenomena such as particle kinematics, heat and mass transfer, and reaction-diffusion. These Baldwinian-PINNs are capable of fast and accurate physics-aware predictions on previously unseen tasks, demonstrating up to several orders of magnitude computation speedup along with an order of magnitude improvement in prediction accuracy relative to recent meta-learned PINNs~\cite{penwarden2023metalearning,liu2022novel}.

The key contributions of this paper are summarized below.
\begin{itemize}
    \item This is the first study to unveil Baldwinian evolution as a compelling route towards discovering neural nets with high capacity to learn diverse physics tasks, thereby advancing generalizability in PINNs.
    \item An instantiation of the Baldwinian evolution framework is proposed through a novel two-stage stochastic programming formulation, wherein the first stage evolves the initial layers of a generalizable PINN model and the second stage trains its final layers to specialize to any new physics task (analogous to lifetime learning).
    \item Comprehensive numerical experimentation and analysis shows that this methodology produces PINNs that effectively learn and predict across multiple PDE problems spanning different physics domains, demonstrating significant advancements in terms of speed and accuracy over models meta-learned by gradient descent.
\end{itemize}

The remainder of the paper is organized as follows. The basic problem setup for a PINN, the notion of generalization over physics tasks, and related work are presented in Section II. The proposed methodology for evolving Baldwinian-PINNs is detailed in Section III. Extensive numerical assessment of the method is carried out in Section IV over a range of linear and nonlinear ODE/PDEs. The paper is concluded in Section V with a discussion of future research directions. 

\section{Preliminaries}
\subsection{Problem setup}\label{subsec:generalproblem}

\subsubsection{Single PINN problem}
For simplicity of exposition, let us consider a problem with single spatial dimension $x$, time dimension $t$, and a single variable of interest $u$. In general, PINNs can learn a mapping between the input variables $(x,t)$ and the output variable $u$ while satisfying specified governing equations representing the physical phenomenon or dynamical process of interest:
\begin{subequations} \label{eq:pde_ibc_eqn}
    \begin{align}
        & \text{PDE:} & \mathcal{N}_\vartheta[u(x,t)] &= h(x,t), & x\in\Omega, t\in(0,T] \label{eq:pde_eqn} \\
        & \text{IC:} & u(x,t=0) &= u_0(x), & x\in\Omega \label{eq:ic_eqn} \\
        & \text{BC:} & \mathcal{B}[u(x,t)] &= g(x,t), & x\in\partial\Omega, t\in(0,T] \label{eq:bc_eqn}
    \end{align}
\end{subequations} 
where the general differential operator $\mathcal{N}_\vartheta[u(x,t)]$ can include linear and/or nonlinear combinations of temporal and spatial derivatives and PDE parameters $\vartheta$, and $h(x,t)$ is an arbitrary source term in the domain $x\in\Omega, t\in(0,T]$. The IC (Eq. \ref{eq:ic_eqn}) specifies the initial state, $u_0(x)$, at time $t=0$, and the BC (Eq. \ref{eq:bc_eqn}) specifies that $\mathcal{B}[u(x,t)]$ equates to $g(x,t)$ at the domain boundary $\partial\Omega$.

Crucially, individual PINN models arrive at an accurate and physics-compliant prediction $u(x,t)$ for a single target scenario by minimizing the discrepancy between Eq. \ref{eq:pde_ibc_eqn} and the model's prediction during training.

\subsubsection{Generalizable neural physics solver}

While most PINN models are trained to solve a specific physics task, there is increasing interest in generalizable neural physics solvers, i.e. models that can be flexibly applied to multiple physics problems once trained. In this context, we can consider a physical phenomena of interest that is represented by a set of training tasks belonging to some underlying task-distribution $p(\boldsymbol{\mathcal{T}})$, e.g., a family of PDEs spanning different PDE parameters $\vartheta$, different ICs $u_0(x)$ and/or different BCs $g(x,t)$. 

Hence, the goal is to discover generalizable PINN models capable of fast, accurate, and physics-aware predictions on unseen scenarios, i.e., any new task from the distribution, $\mathcal{T}_i \backsim p(\boldsymbol{\mathcal{T}})$, by learning the underlying governing physics. In the context of meta-learning, the learning objective is to use training tasks from $p(\boldsymbol{\mathcal{T}})$ to find network initializations that are most amenable to a quick and accurate solution for the PINN loss, thereby accelerating the solution of multiple related physics problems at test time and providing a potential route to a generalizable neural physics solver.

\subsection{Related Work}\label{subsec:relatedwork}

PINN models are usually trained to solve a single, specific physics task. However, recent studies on meta-learning of PINNs to solve different physics scenarios as separate tasks have emerged, although no work has been reported from the neuroevolution perspective to our knowledge.

Most of the meta-learning PINN approaches reported in literature use weight interpolation as the basic framework. The simplest instantiation of this approach is to first train independent PINN models for each task and then interpolate across model weights for the new task~\cite{penwarden2023metalearning, chen2024gpt}. Several interpolation methods such as the Gaussian Process (GP) and Radial Basis Function (RBF) have been studied and shown to improve physics-informed learning on new tasks~\cite{penwarden2023metalearning}. In other studies, interpolation is learned through the use of hypernetworks, whereby the hypernetworks and PINN are trained simultaneously from all the tasks~\cite{cho2024hypernetwork, toloubidokhti2023dats}. Other variants include encoding tasks as latent variables and passing them into the input layer of the PINN model~\cite{huang2022meta}.

However, there are two major drawbacks to such weight interpolation frameworks. Firstly, these methods rely heavily on task parameterization to perform interpolation and operate under the assumption of smoothness. This means that all the train tasks (and the new task) must adhere to this parameterization requirement. In addition, these methods do not capitalize on the principal characteristic of PINN, which is the potential for physics-informed learning (retraining or fine-tuning) for a new task, in their meta-learning formulation. The data-free nature of physics-informed fine-tuning is not exploited during the meta-learning phase, given that there is no guiding principle on how the fine-tuning towards any new task should be performed given the interpolated weights. 


\begin{figure*}[h]%
\centering
\includegraphics[width=1.0\textwidth]{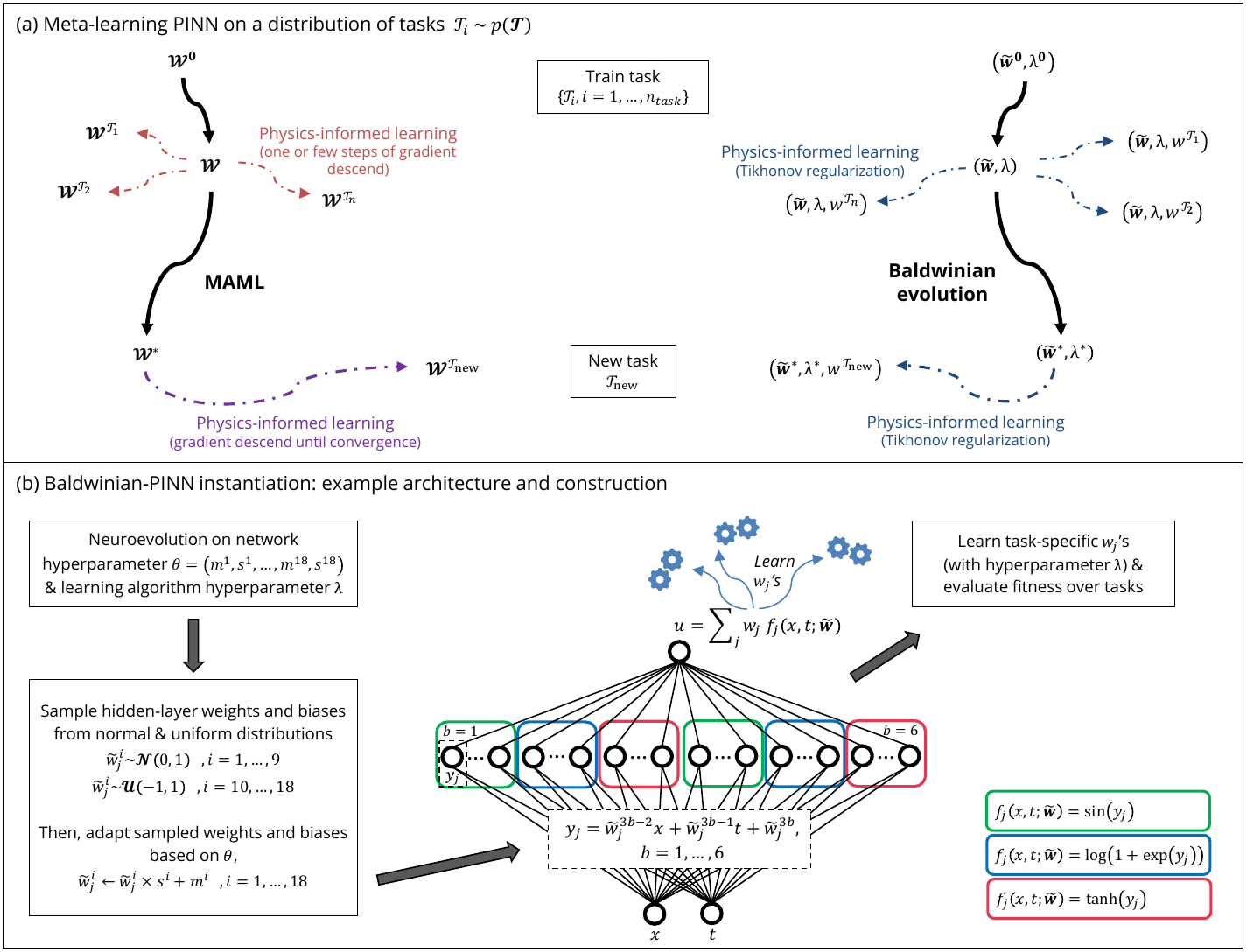}
\vspace{-0.6cm}
\caption{ (a) Meta-learning PINN with Baldwinian neuroevolution (right) versus MAML (left). In MAML, the initial weights $\vb*{W}$ are learned using gradient-based method, such that they can be quickly fine-tuned (physics-informed learning) on new tasks. Although the task-specific fine-tuning is limited to one or a few gradient descent updates during training, such amount of fine-tuning is usually insufficient for a PINN at test time. In Baldwinian neuroevolution, the weight distribution in the pre-final nonlinear hidden layers $\vb*{\tilde{w}}$ and learning hyperparameters $\lambda$ are jointly evolved. The task-specific physics-informed learning is performed on the final layer weights $w$ (segregated from $\vb*{\tilde{w}}$) with a 1-step Tikhonov regularization operation at both training and test time (for linear ODE/PDEs). (b) Schematic of Baldwinian-PINNs architecture used in present study and procedure to obtain nonlinear hidden layers’ weights $\vb*{\tilde{w}}$’s from the evolved network hyperparameter $\theta$ for learning task-specific outputs.}\label{fig:architecture}
\end{figure*}


The model-agnostic meta-learning (MAML)~\cite{finn2017model} framework can theoretically overcome the limitations of weight interpolation methods. MAML and its variant, Reptile~\cite{nichol2018reptile}, aim to learn an optimal weight initialization that can be quickly fine-tuned on new tasks without the need for interpolation~\cite{liu2022novel}. An illustration of MAML for PINN is shown in Figure~\ref{fig:architecture}(a). To reduce the computation cost of meta-learning, task-specific physics-informed learning is limited to one or a few gradient descent updates from the initialization. This essentially assumes that the model performance after one step (or a few steps of) gradient descent update from the initialization is already indicative of the learning performance. While this assumption may be appropriate in data-driven models, PINN models are much more challenging to optimize. Hence, they require more thorough training (e.g., a large number of updates, even with the use of 2nd-order gradient descent methods) to achieve a good convergence. Moreover, gradient-based MAML methods are prone to getting stuck in a local minimum and struggle to find a good initialization for a diverse set of tasks, as the gradients can be noisy or even deceptive, i.e., requiring going against it to reach the optimum. Hence, recent studies have shown poor MAML results, potentially due to non-convergence during meta-learning or insufficient physics-informed fine-tuning performed at test time~\cite{penwarden2023metalearning, cho2024hypernetwork}.

\section{Proposed Methodology}\label{sec:methods}

We address the above limitations with a meta-learning PINN framework based on Baldwinian neuroevolution. A novel evolutionary algorithm is crafted to jointly optimize the weight distribution in the network’s hidden layers alongside other hyperparameters that are essential for achieving optimal performance on downstream physics-informed learning tasks. We limit our task-specific learning to the final layer, allowing us to solve linear problems with a 1-step Tikhonov regularization operation (or a few steps for nonlinear problems) during both the meta-learning process and test time, in a consistent manner. Importantly, once given the nonlinear hidden layers' weights and learning hyperparameters, the Tikhonov regularization is exact and fast (does not require further gradient descent updates). An illustration of the Baldwinian evolution of PINNs is shown in Figure~\ref{fig:architecture}(a). 

\subsection{The Baldwin effect as a two-stage stochastic program}\label{subsec:twostageproblem}
Without loss of architectural generality, Baldwinian-PINNs can be assigned the form of a multilayer perceptron (MLP) with proven representation capacity \cite{hornik1989multilayer, zhang2012universal}, thereby ensuring the ability to learn a diversity of tasks. Its output $u(x,t)$ can be written as:
\begin{equation}
    u(x,t) = \sum_j w_j f_j(x,t;\vb*{\tilde{w}}) \label{eq:uexpansion}
\end{equation}
where $\vb*{w} = [\dots w_j \dots]^T$ are the final layer weights, and $f_j(x,t;\vb*{\tilde{w}})$’s represent nonlinear projections of the input variables with the hidden layers’ weights $\vb*{\tilde{w}}$. The connection strengths $\vb*{\tilde{w}}$ are typically learned during PINN training. However, in the proposed implementation of Baldwinian-PINNs as detailed below, they are drawn from a probability distribution defined by a dimensionally reduced set of network hyperparameters $\theta$---see example in Section~\ref{subsec:architecture} and Figure. \ref{fig:architecture}(b)---that are assigned \textit{at birth} as per Baldwinian evolution.

The model's reaction to a given environment\textemdash i.e., finding the best set of $w_j$’s such that the model's output satisfies Eq. \ref{eq:pde_ibc_eqn} for a specific physics task $\mathcal{T}_i$\textemdash is the focus of a typical PINN described in Section~\ref{subsec:generalproblem}. This is reduced to a physics-informed least-squares problem in this work:
\begin{equation}
    \vb*{w^{*}} = \argmin_{\vb*{w}} \ (\vb{A}_{\vb*{\tilde{w}}}^{\mathcal{T}_i} \vb*{w} - \vb{b}^{\mathcal{T}_i})^{\vb{T}}(\vb{A}_{\vb*{\tilde{w}}}^{\mathcal{T}_i} \vb*{w} - \vb{b}^{\mathcal{T}_i}) + \lambda \vb*{w}^{\vb{T}}\vb*{w} \label{eq:stage2problem}
\end{equation}
where $\vb{A}_{\vb*{\tilde{w}}}^{\mathcal{T}_i}\vb*{w}$ is obtained by substituting the model's output into the left hand side of Eq. \ref{eq:pde_ibc_eqn} for a given set of collocation points, and $\vb{b}^{\mathcal{T}_i}$ represents the corresponding right hand side of Eq. \ref{eq:pde_ibc_eqn}. Eq. \ref{eq:stage2problem} yields a closed-form solution $\vb*{w^{*}}$ when using the Moore-Penrose generalized inverse, permitting extremely fast learning (order of milli-seconds in our experiments) for physics-compliant prediction. Similar least-squares formulations studied in the literature have been shown to be competitive with, or faster than, widely-used numerical solvers such as the finite element method \cite{dong2021local, dong2022computing}. The learning hyperparameter $\lambda \geq 0$ reduces the $L^2$-norm of the least-squares solution, thereby improving the solution numerically. Detailed derivations, including information on the construction of matrix $\vb{A}_{\vb*{\tilde{w}}}^{\mathcal{T}_i}$ and vector $\vb{b}^{\mathcal{T}_i}$, are provided in Section~\ref{subsec:frpinn}.

Going beyond specialization to a single physics task by means of Eq. \ref{eq:stage2problem}, we formulate the search for Baldwinian-PINN models that can generalize to a whole family of PDEs as the following two-stage stochastic optimization problem~\cite{bakker2020structuring}: 

\begin{align}
\nonumber
    \min_{\theta,\lambda} \E_{\mathcal{T}_i \backsim p(\boldsymbol{\mathcal{T}})} \E_{\vb*{\tilde{w}} \backsim p_{\theta}(\vb*{\tilde{w}})} &[ \kappa_{LSE} \ l_{LSE}(\vb*{w^{*}}) + \kappa_{MSE} \ l_{MSE}(\vb*{w^{*}}) ] \quad \\ 
    & \text{subject to } \kappa_{LSE} \geq 0, \ \kappa_{MSE} \geq 0
    \label{eq:stage1problem}
\end{align}
where $\vb*{w^{*}}$ is the solution of the second stage problem defined earlier in Eq. \ref{eq:stage2problem}, which provides optimal final layer weights that allow the model to specialize to any realization of task $\mathcal{T}_i \backsim p(\boldsymbol{\mathcal{T}})$ for the given network's $\vb*{\tilde{w}} \backsim p_{\theta}(\vb*{\tilde{w}})$.

The first stage optimization objective is defined by the weighted sum of the physics learning proficiency, i.e., sum of squared residuals or least-squares error (LSE),
\begin{equation}
    l_{LSE}(\vb*{w^{*}}) = (\vb{A}_{\vb*{\tilde{w}}}^{\mathcal{T}_i} \vb*{w^{*}} - \vb{b}^{\mathcal{T}_i})^{\vb{T}}(\vb{A}_{\vb*{\tilde{w}}}^{\mathcal{T}_i} \vb*{w^{*}} - \vb{b}^{\mathcal{T}_i}) \label{eq:lse}
\end{equation}
and the actual predictive performance, i.e., mean squared error (MSE),
\begin{equation}
    l_{MSE}(\vb*{w^{*}}) = \frac{1}{n} \sum_{s=1}^n \ (u_s^{label} - \sum_j w_j^{*} f_j(x_s,t_s;\vb*{\tilde{w}}))^2 \label{eq:mse}
\end{equation}
given labelled data $u_s^{label}, s=1,...,n$, over the task distribution $p(\boldsymbol{\mathcal{T}})$ and the network connections' distribution $p_{\theta}(\vb*{\tilde{w}})$. 

\emph{In what follows, we present a novel procedure for solving the two-stage stochastic program via an evolutionary algorithm. The hyperparameters $\theta,\lambda$ are evolved from one generation to the next, with their fitness evaluations for the first stage problem conditioned on lifetime learning to obtain the optimal $\vb*{w^{*}}$ from the second stage problem. Note that while the outcome of lifetime learning influences the selection pressure acting on the evolving hyperparameters, it does not directly alter the genetic makeup of the hyperparameters within a generation. As such, our overall method exhibits a clear connection with the evolutionary principles of Baldwinism}.

\subsection{Baldwinian neuroevolution}\label{subsec:algorithm}

The Baldwinian neuroevolution procedure to solve the two-stage stochastic programming problem defined in Section~\ref{subsec:twostageproblem} is described in Algorithm~\ref{algo:baldwinianlearning} and Algorithm~\ref{algo:lifetimelearning}. Recall from Figure. \ref{fig:architecture}(b) that a Baldwinian-PINN is represented by $(\theta,\lambda)$, i.e., distribution of network weights and biases and lifetime learning hyperparameters, during neuroevolution.


\begin{algorithm}
\caption{Baldwinian neuroevolution of PINNs}\label{algo:baldwinianlearning}
\begin{flushleft}
        \textbf{INPUT:} training tasks distribution, $p(\boldsymbol{\mathcal{T}})$\\
        \textbf{OUTPUT:} best solution found (i.e., Baldwinian-PINN model's weights and biases in nonlinear hidden-layers and lifetime learning hyperparameters $(\theta,\lambda)$)
\end{flushleft}
\begin{algorithmic}[1]
\Require $\boldsymbol{\mathcal{F}}$: procedure to return fitness based on lifetime learning performance of individual given a batch of tasks (described in detail in Algorithm~\ref{algo:lifetimelearning})
\State Initialize population $\boldsymbol{\mathcal{P}}$
\While{not done}
        \State Sample a new batch of offspring $(\theta^g,\lambda^g), g=1,...,n_{pop}$ from the population $\boldsymbol{\mathcal{P}}$
        \State Sample batch of tasks $\mathcal{T}_i \backsim p(\boldsymbol{\mathcal{T}}), i=1,...,n_{task}$
        \For{all $(\theta^g,\lambda^g)$}
            \State $f^g = \boldsymbol{\mathcal{F}}(\theta^g,\lambda^g, \{\mathcal{T}_i, i=1,...,n_{task}\})$
        \EndFor
        \State Update population $\boldsymbol{\mathcal{P}}$ based on fitness of individuals: $\{(\theta^g,\lambda^g,f^g), g=1,....,n_{pop}\}$
\EndWhile
\State Return best individual $(\theta,\lambda)$ found in population $\boldsymbol{\mathcal{P}}$
\end{algorithmic}
\end{algorithm}

\begin{algorithm}
\caption{Baldwinian-PINN lifetime learning and fitness calculation $\boldsymbol{\mathcal{F}}$}\label{algo:lifetimelearning}
\begin{flushleft}
        \textbf{INPUT:} network and lifetime learning hyperparameters $(\theta, \lambda)$, batch of tasks $\mathcal{T}_i, i=1,...,n_{task}$\\
        \textbf{OUTPUT:} fitness $f$
\end{flushleft}
\begin{algorithmic}[1]
\Require $\boldsymbol{\mathcal{G}}$: procedure to sample weights and biases in nonlinear hidden-layers (exemplified in Section~\ref{subsec:architecture} and Figure. \ref{fig:architecture}(b))
\Require $\boldsymbol{\mathcal{C}}$: procedure to construct least squares problem on a set of collocation points based on underlying physics (PDEs, BCs, ICs) of the task (described in detail in Section~\ref{subsec:frpinn})
\State Sample hidden layers’ weights and biases $\vb*{\tilde{w}} = \boldsymbol{\mathcal{G}}(\theta)$ 
\For{all $\mathcal{T}_i$}
    \State Construct least squares matrix and vector:
    \Statex \hspace{\algorithmicindent}$(\vb{A}, \vb{b}) = \boldsymbol{\mathcal{C}}(\mathcal{T}_i, \vb*{\tilde{w}})$
    \State Compute least squares solution as per Eq.~\ref{eq:pseudoinverse}:
    \Statex \hspace{\algorithmicindent}$\vb*{w^{*}} = (\lambda I + \vb{A}^T \vb{A})^{-1} \vb{A} \vb{b}$
    \State Compute least squares error (LSE) as per Eq.~\ref{eq:lse}:
    \Statex \hspace{\algorithmicindent}$l_{LSE}^{\mathcal{T}_i} = (\vb{A} \vb*{w^{*}} - \vb{b})^{\vb{T}}(\vb{A} \vb*{w^{*}} - \vb{b})$ 
    \State Compute mean squared error (MSE) as per Eq.~\ref{eq:mse}:
    \Statex \hspace{\algorithmicindent}$l_{MSE}^{\mathcal{T}_i} = \frac{1}{n} \sum_s^n (u_s^{label} - \sum_j w_j^* f_j(x_s,t_s;\vb*{\tilde{w}}_j))^2$
\EndFor
\State Compute overall fitness: $f = - (\kappa_{LSE} \ \sum_{\mathcal{T}_i} l_{LSE}^{\mathcal{T}_i} + \kappa_{MSE} \ \sum_{\mathcal{T}_i} l_{MSE}^{\mathcal{T}_i})$
\end{algorithmic}
\end{algorithm}

The Baldwinian neuroevolution procedure described in Algorithm~\ref{algo:baldwinianlearning} is generic for evolutionary optimization methods. The algorithm initializes a population $\boldsymbol{\mathcal{P}}$ of Baldwinian-PINN models given by different weights and biases in the nonlinear hidden-layers and lifetime learning hyperparameters $(\theta,\lambda)$. For a probabilistic model-based evolutionary algorithm, $\boldsymbol{\mathcal{P}}$ is commonly represented by a \textit{search distribution}. In each generation, $n_{pop}$ new offspring individuals are sampled from the search distribution (probabilistic model-based EAs) or through crossover/mutation (traditional EAs), and their fitness are evaluated for a batch of tasks randomly sampled from the training task distribution $p(\boldsymbol{\mathcal{T}})$. The fitness evaluation procedure $\boldsymbol{\mathcal{F}}$ gives the lifetime learning outcome $f^g$ of these offspring individuals $(\theta^g,\lambda^g), g=1,...,n_{pop}$ for the given tasks. In line with the essence of Baldwinism, the lifetime learning procedure (described in detail in Algorithm~\ref{algo:lifetimelearning}) does not alter the genetic makeup $(\theta,\lambda)$ of the individuals. The outcome of the lifetime learning procedure specifies the fitness which creates the selection pressure influencing the evolution of the population, but is not directly inherited by the next population (unlike Lamarckian evolution). The Baldwinian neuroevolution algorithm iteratively adapts $\boldsymbol{\mathcal{P}}$ towards offspring with better fitness until the convergence criteria, e.g., a pre-determined number of generations or fitness value (trade-off between computation resource and convergence), is met.

Given an individual $(\theta,\lambda)$ sampled from the search distribution and a batch of tasks sampled from the training distribution, $\mathcal{T}_i \backsim p(\boldsymbol{\mathcal{T}}), i=1,...,n_{task}$, the procedure $\boldsymbol{\mathcal{F}}$ to return fitness is detailed in Algorithm~\ref{algo:lifetimelearning}. It starts with a procedure $\boldsymbol{\mathcal{G}}$ to populate nonlinear hidden-layers’ weights and biases $\vb*{\tilde{w}}$ of a Baldwinian-PINN from the sampled individual $\theta$. In the present study, Baldwinian-PINNs are designed to have $\vb*{\tilde{w}}$ fixed \textit{at birth} to random values drawn from a probability distribution defined by a dimensionally reduced set of network hyperparameters $\theta$, akin to a randomized neural networks setup \cite{suganthan2021origins, bakker2020structuring}. This procedure is exemplified in Section~\ref{subsec:architecture}. Then, lifetime learning of the Baldwinian-PINN is performed to obtain the optimal network weights $\vb*{w^{*}}$ in the final layer, for each of the sampled tasks $\mathcal{T}_i$. It involves procedure $\boldsymbol{\mathcal{C}}$ to construct least squares problem, i.e., matrix and vector $(\vb{A}, \vb{b})$, on a fixed set of collocation points based on underlying physics (PDEs, BCs, ICs) of the task. This procedure is detailed in Section~\ref{subsec:frpinn}. We choose the collocation points to coincide with the location of labelled data for MSE computation, although this is not a prerequisite. The overall fitness $f$ is computed by aggregating the LSE and MSE based on lifetime learning outcomes for all the sampled tasks. 

In the spirit of the Baldwin effect, the task-specific final layer $\vb*{w^{*}}$ is the outcome of lifetime learning and not directly inherited by the next generation of offspring~\cite{gupta2018memetic}; only the hyperparameters $(\theta,\lambda)$ are subjected to evolutionary variation and inheritance. 

\subsection{Baldwinian-PINN lifetime learning procedure}\label{subsec:frpinn}

The lifetime learning procedure of the Baldwinian-PINNs occurs only in the linear final layer of the network, i.e., finding the best set of $w_j$’s such that the output $u(x,t) = \sum_j w_j f_j(x,t;\vb*{\tilde{w}})$ satisfies the governing equations in Eq. \ref{eq:pde_ibc_eqn} for a specific task. Given a set of collocation points $(x_i^{pde},t_i^{pde}), i=1,...,n_{pde}, (x_i^{ic},0), i=1,...,n_{ic}, (x_i^{bc},t_i^{bc}), i=1,...,n_{bc}$ sampled from the respective domain, the following system of equations can be formed:  

\begin{align}
\nonumber
    \scalebox{0.85}{$ 
    \left[
    \begin{array}{ccc}
        \dots & \mathcal{N}_\vartheta[f_j(x_{1}^{pde},t_{1}^{pde};\vb*{\tilde{w}})] & \dots \\ 
         & \vdots &    \\
        \dots & \mathcal{N}_\vartheta[f_j(x_{n_{pde}}^{pde},t_{n_{pde}}^{pde};\vb*{\tilde{w}})] & \dots \\ \\
        \dots & f_j(x_{1}^{ic},0;\vb*{\tilde{w}}_j) & \dots \\ 
         & \vdots &    \\
        \dots & f_j(x_{n_{ic}}^{ic},0;\vb*{\tilde{w}}) & \dots \\ \\      
        \dots & \mathcal{B}[f_j(x_{1}^{bc},t_{1}^{bc};\vb*{\tilde{w}})] & \dots \\ 
         & \vdots &    \\
        \dots & \mathcal{B}[f_j(x_{n_{bc}}^{bc},t_{n_{bc}}^{bc};\vb*{\tilde{w}})] & \dots \\
    \end{array}
    \right] 
    \left[
    \begin{array}{c}
        \vdots \\
        w_j    \\
        \vdots \\
    \end{array}
    \right]    
    =
    \left[
    \begin{array}{c}
        h(x_{1}^{pde},t_{1}^{pde})  \\ 
        \vdots  \\ 
        h(x_{n_{pde}}^{pde},t_{n_{pde}}^{pde})  \\ \\
        u_{0}(x_{1}^{ic})  \\ 
        \vdots  \\ 
        u_{0}(x_{n_{ic}}^{ic})  \\ \\
        g(x_{1}^{bc},t_{1}^{bc})  \\ 
        \vdots  \\ 
        g(x_{n_{bc}}^{bc},t_{n_{bc}}^{bc})  \\ 
    \end{array}
    \right]
    $} 
\end{align}
\begin{align}
    \vb{A} \vb*{w} &= \vb{b} \label{eq:leastsquares}  
\end{align}

Note that the derivatives required to construct $\vb{A}$ can be easily computed by automatic differentiation \cite{baydin2018automatic}. The best-fit solution to the above system of linear equations with unknown $\vb*{w} = [\dots w_j \dots]^T$ can be obtained by means of the Tikhonov regularization:
\begin{equation} \label{eq:pseudoinverse}
        \vb*{w^{*}} = (\lambda I + \vb{A}^T \vb{A})^{-1} \vb{A}^T \vb{b} 
\end{equation}
where the learning hyperparameter $\lambda \geq 0$ reduces the $L^2$-norm of the least-squares solution, thereby improving the solution numerically. Such a least-squares formulation yields a closed-form result in a single computation step for any linear PDE (i.e., the governing equations in Eq. \ref{eq:pde_ibc_eqn} are all linear with respect to $u$) which encapsulates a wide class of physics phenomena in the natural world.


For nonlinear PDEs, iterative methods can be used for arriving at optimized $w_j$’s in Baldwinian-PINNs (detailed in Suppl.~\ref{subsec:lagging}). Hence, the Baldwinian-PINNs' lifetime learning is broadly applicable to \textbf{both} linear and nonlinear PDEs, with the psuedoinverse formulation permitting extremely fast computation for physics-compliant prediction. It is worth emphasizing that each Baldwinian-PINN undergoes lifetime learning in order to produce accurate, physics-compliant predictions on a single, new physics task, hence, the fast nature of this procedure permits flexible and rapid prediction for each new task on-demand. In addition, there is no requirement on prior labelled data for any new task as the learning can be entirely physics-informed (as per conventional PINNs).

\subsection{Implementation details}\label{subsec:algorithm_detail}

\subsubsection{Baldwinian-PINN architecture}\label{subsec:architecture}

In the present study, Baldwinian-PINNs are designed to have weights and biases in the nonlinear hidden-layers $\vb*{\tilde{w}}$ fixed to random values \textit{at birth}. They are akin to a randomized neural networks setup \cite{suganthan2021origins, bakker2020structuring} and drawn from a probability distribution defined by a dimensionally reduced set of network hyperparameters $\theta$. Both normal and uniform distributions for the weights and biases are possible for the randomized PINNs \cite{dong2021local,dong2022computing}. Similarly, smooth nonlinear activations such as \textit{sin}, \textit{softplus}, and \textit{tanh} are common in PINN literature and have their own merits. For greater flexibility, we apply both distributions for setting weights and biases and all three activations to different hidden layer blocks in the Baldwinian-PINN models.

Our base Baldwinian-PINN model is depicted in Figure~\ref{fig:architecture}(b). It's hidden layer architecture is segmented into $3 \times 2 = 6$ unique blocks, with the weights and biases of the first 3 blocks sampled from normal distributions and the weights and biases of the other 3 blocks sampled from uniform distributions. Each block has a fixed number of neurons ($n_{neuron}$ = 150 or 200). Assuming 2 input variables $(x,t)$, we can write the output $f_j$ for all the neurons $j=1,...,n_{neuron}$ in a nonlinear hidden-layer block as:
\begin{subequations} \label{eq:hiddenneuron}
    \begin{align}
        &y_j = \tilde{w}_j^{3b-2} x + \tilde{w}_j^{3b-1} t + \tilde{w}_j^{3b} \label{eq:hiddenneuron_1} \\
        &f_j(x,t;\vb*{\tilde{w}}_j) = \varphi^b(y_j) \label{eq:hiddenneuron_2}
    \end{align}
\end{subequations} 
for each of the blocks $b=1,...,6$, where the activation $\varphi^b$ can be \textit{sin} ($b=1,4$), \textit{softplus} ($b=2,5$), or \textit{tanh} ($b=3,6$). The $\tilde{w}_j^{i}$’s are weights and biases with their own distributional mean $m^i$ and spread $s^i$, $i=1,...,18$. Their values can be obtained by the following sampling procedure:
\begin{subequations} \label{eq:populateweights}
    \begin{align}
        & \tilde{w}_j^{i} \sim \mathcal{N}(0,1) \quad \text{or} \quad \tilde{w}_j^{i} \sim \mathcal{U}(-1,1) \\
        & \tilde{w}_j^{i} \gets \tilde{w}_j^{i} \times s^{i} + m^{i}  
    \end{align}
\end{subequations} 

Given the Baldwinian-PINNs’ configuration, a set of network hyperparameters $\theta=(m^{1},s^{1},...,m^{18},s^{18})$ control the distributional mean and spread of the weights and biases in different blocks.
Since the neuroevolution only searches the distribution parameters for groups of weights instead of evolving each individual weight in the nonlinear hidden layer, the reduced dimensionality can be even more effectively searched by today's evolutionary algorithms.

While the above description is for a single hidden layer, we note that the Baldwinian-PINN framework is not restricted to a single hidden layer. We include additional examples in Suppl.~\ref{subsec:additional_evo_depth} to show that the methodology also works for deeper neural architectures. 

\subsubsection{Evolutionary algorithm}\label{subsec:algorithm_description}

In the majority of this study, we employ the covariance matrix adaptation evolution strategy (CMA-ES) \cite{hansen2016cma} for evolving $(\theta,\lambda)$, although results in Suppl.~\ref{subsec:additional_evo_depth} show the extensibility of this framework to other neuroevolution algorithms. As an instantiation of information-geometric optimization algorithms \cite{ollivier2017information}, CMA-ES represents the population of $(\theta,\lambda)$ in $\boldsymbol{\mathcal{P}}$ using a multivariate normal search distribution, initialized with zero mean and standard deviation (std.) as tuning hyperparameter. It iteratively adapts the search distribution based on the rank-based fitness landscape until the convergence criteria is met. Our experiments show that the performance of Baldwinian neuroevolution is robust across a range of CMA-ES hyperparameters such as population size and initial standard deviation (std.) of search distribution, and the number of tasks sampled for fitness evaluation per iteration. Hence, a robust setting that shows good convergence in fitness across different types of problems is chosen based on initial experiments.  

The network weights and biases can take any value from $(-\infty$, $\infty)$, hence there is no restriction to the continuous search space of $\theta$ representing their distributional means and spreads. The learning hyperparameter $\lambda \geq 0$ can be evolved in continuous search space and its absolute value is then used for computing the least-squares solution. In our implementation with CMA-ES, $(\theta,\lambda)$ share the same initial standard deviation, and we scale the learning hyperparameter by a factor, i.e., $\lambda \gets 1\mathrm{e}{-4} \times \text{abs}(\lambda)$, to improve the performance of the Tikhonov regularization.

In our preliminary experiments, we found it helpful to have the $l_{MSE}$ component in the optimization objective even though the learning of future test tasks remains solely physics-informed. We set $\kappa_{LSE} = \kappa_{MSE} = 1$ as default for the computation of overall fitness unless there is a huge difference in magnitude between $l_{LSE}$ and $l_{MSE}$ during Baldwinian neuroevolution. In addition, we can multiply the BC/IC rows in both $\vb{A}$ and $\vb{b}$ to re-balance the importance between PDE and BC/IC errors in the least-squares solution.

As this study focuses on the paradigm of Baldwinian neuroevolution as a pathway towards generalizable neural physics solvers, other combinations of evolutionary optimization or lifetime learning algorithms may be used. One key advantage of evolutionary optimization is that the fitness evaluations (population size $n_{pop}$ $\times$ number of random tasks $n_{task}$) required each iteration can be easily parallelized across multiple GPUs to fully harvest any hardware advantage. In particular, we utilized the JAX framework to harness previously reported performance improvements for automatic differentiation and linear algebra operations \cite{bradbury2018jax,evojax2022}. The experimental study is performed on a workstation with an Intel Xeon W-2275 Processor and 2 NVIDIA GeForce RTX 3090 GPUs.

\section{Experimental Studies}\label{sec:results}


\begin{figure*}[h]%
\centering
\includegraphics[width=1.0\textwidth]{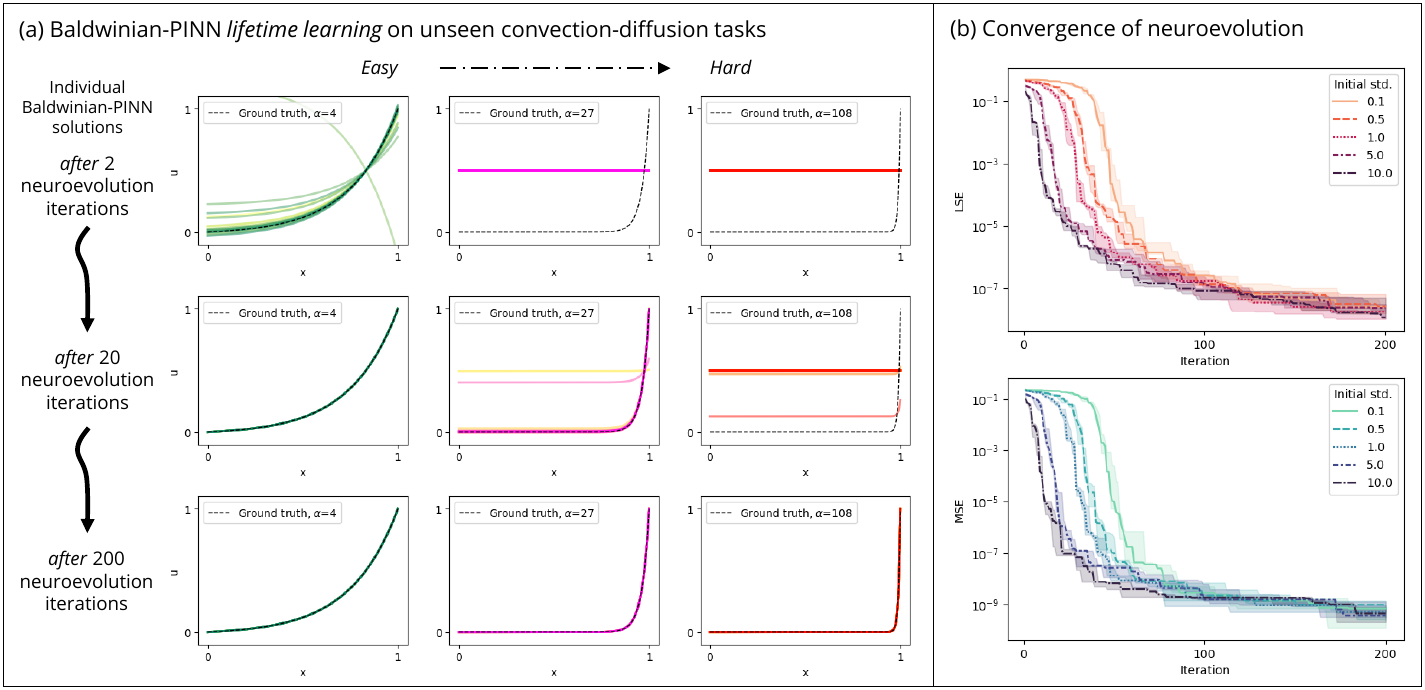}
\vspace{-0.6cm}
\caption{(a) Solution of 20 individual Baldwinian-PINN models sampled from the CMA-ES search distribution (initial std. = 1), for unseen convection-diffusion tasks $\alpha = \{4,27,108\}$. The task is more challenging with increasing $\alpha$. Baldwinian neuroevolution is effective for evolving good Baldwinian-PINN models which can generalize across different difficulties. (b) Baldwinian neuroevolution demonstrates effective LSE and MSE convergences on convection-diffusion problem, for different CMA-ES initial std. values (best std. = 1). The bold lines indicate their median convergence path from 5 individual runs, and the shaded areas indicate their interquartile ranges.}\label{fig:cd}
\end{figure*}


We examine the efficacy of Baldwinian neuroevolution for learning physics (Algorithm~\ref{algo:baldwinianlearning} and Algorithm~\ref{algo:lifetimelearning} in Section~\ref{subsec:algorithm}) as formalized in a two-stage stochastic programming problem in Section~\ref{subsec:twostageproblem}. Several ODE/PDE problems which are representative of real-world phenomena are used to demonstrate Baldwinian neuroevolution for physics in the following sections. Table~\ref{tab:summary} summarizes neuroevolution configurations and Baldwinian-PINN lifetime learning performance on their respective test tasks, including computational time for both neuroevolution and lifetime learning phases.

\subsection{Learning to solve and generalize linear ODE/PDEs}\label{subsec:linearproblems}

\subsubsection{Convection-diffusion}\label{subsec:linearode}

The steady-state convection-diffusion equation is a ubiquitous physics model that describes the final distribution of a scalar quantity (e.g. mass, energy, or temperature) in the presence of convective transport and diffusion~\cite{stynes2005steady}. Solutions to this physics are key to characterization and design of many systems, including microfluidic chip cooling in electronics~\cite{van2020co}. The 1D equation is defined as:
\begin{equation}
    \text{(Problem 1)} \quad\quad \alpha \frac{du}{dx} - \frac{d^2u}{dx^2} = 0 \quad , x\in[0,1] \label{eq:cd}
\end{equation}
subject to BCs $u (x=0) = 0$; $u (x=1) = 1$. These real-world problems have characteristic physics that vary with non-dimensional constants such as the Peclet number $Pe$ (ratio of convection to diffusion)~\cite{Patankar80}. Hence, it is helpful to learn a PINN model that can return $u(x)$ for a diverse range of $Pe$-related problems (determined by $\alpha$ here). The training tasks consist of $\alpha = \{5,10,...,100\}$, encompassing both smoother output patterns at lower $\alpha$ and very high gradient patterns at higher $\alpha$, with the latter being challenging for PINNs to learn by both stochastic gradient descent (SGD) \cite{wong2021can,wong2022learning} and classical numerical methods~\cite{gupta2013numerical}.

The predictive performance of the learned model is evaluated for an unseen range of test tasks, i.e., $\alpha = \{1,2,...,110\}$. The efficacy of Baldwinian neuroevolution is demonstrated in Figure~\ref{fig:cd}, with the successful evolution of Baldwinian-PINNs which can learn an extremely accurate solution on new test tasks in milli-seconds. The learned solutions can achieve an average MSE of 5.8$\mathrm{e}{-9}$ {\tiny$\pm$9.9$\mathrm{e}{-9}$} ($n=110$ tasks $\times$ 5 individual runs) after 200 neuroevolution iterations.

\subsubsection{Family of linear PDEs (convection, diffusion, and dispersion)}\label{subsec:linearpde}


\begin{figure*}[h]%
\centering
\includegraphics[width=1.0\textwidth]{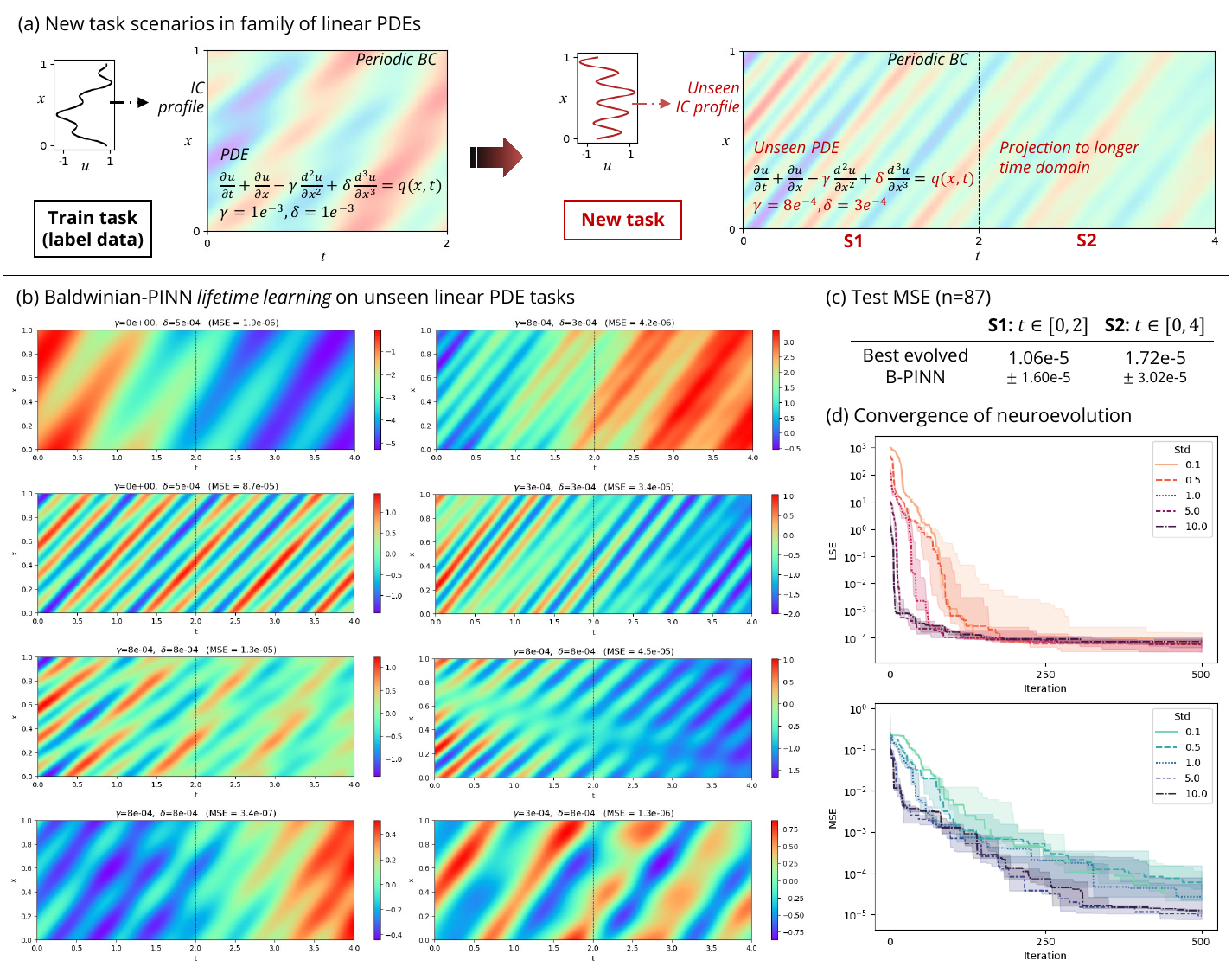}
\vspace{-0.6cm}
\caption{(a) Schematic to illustrate new tasks arising from family of linear PDEs problem: \textbf{S1} change to new PDE and IC profile for $t\in[0,2]$ (same time domain as train tasks), and \textbf{S2} projection to longer time domain $t\in[0,4]$. (b) Solution for unseen linear PDE tasks obtained by best evolved Baldwinian-PINN sampled from the center of CMA-ES search distribution after 500 iterations with initial std. = 5 (they are visually indistinguishable from the ground truth). (c) The mean MSE over $n=87$ test tasks for 2 test scenarios described in (a) are below $5\mathrm{e}{-5}$. (d) Baldwinian neuroevolution demonstrates effective LSE and MSE convergences for different CMA-ES initial std. values, with a superior performance given by std. = 5 and 10. The bold lines indicate their median convergence path from 5 individual runs, and the shaded areas indicate their interquartile ranges.}\label{fig:linear}
\end{figure*}


Next, we extend Baldwinian-PINNs to a family of linear PDEs. This PDE family is a further generalization of the convection-diffusion equation:
\begin{subequations} \label{eq:linear}
    \begin{multline} 
        \text{(Problem 2)} \quad \frac{du}{dt} + \alpha \frac{du}{dx} - \gamma \frac{d^2u}{dx^2} + \delta \frac{d^3u}{dx^3} = q(x,t), \\
         x \in [0,1], \quad t \in (0,2] 
        \label{eq:linearpde}
    \end{multline}
    \begin{multline} 
        q(x,t) = \sum_{j=1}^{J} A_j \sin\left(\omega_j t + \frac{2\pi l_j x}{L} + \varphi_j\right)
        \label{eq:linearpdesource}
    \end{multline}
\end{subequations}
with IC $u(x,0)=q(x,0)$ and periodic BC $u(0,t)=u(1,t)$. Eq. \ref{eq:linear} models the time evolution of a scalar quantity in the presence of physics phenomena such as convection ($\frac{du}{dx}$ component), diffusion ($\frac{d^2u}{dx^2}$ component), and dispersion ($\frac{d^3u}{dx^3}$ component), and a rich diversity of dynamical processes can be generated from different PDE and IC combinations \cite{bar2019learning, brandstetter2022message}. $q(x,t)$ is a source term, with different $q(x,t=0)$ being the corresponding IC profiles. We consider the following PDE scenarios: $\alpha = 1$, $\gamma = \{0, 5\mathrm{e}{-4}, 1\mathrm{e}{-3}\}$, $\delta = \{0, 5\mathrm{e}{-4}, 1\mathrm{e}{-3}\}$. The ratio between $\alpha$, $\gamma$, and $\delta$ determine non-dimensional constants (e.g. $Pe$), and consequently, the systems' characteristic physics. $q(x,t)$ comprises scenarios with $J = 5$, $L = 6$ and coefficients sampled uniformly from $A_j\in[-0.8,0.8]$, $\omega_j\in[-2,2]$, $l_j\in[0,1,2,3,4]$, $\varphi_j\in[-\pi,\pi]$. The training set comprises 108 tasks with different PDE and IC combinations.

The effective generalization of a Baldwinian-PINN to an entire PDE family with diverse output patterns is demonstrated on 2 task scenarios: \textbf{S1} shows successful learning of $u(x,t)$ for $t\in[0,2]$ on unseen set of PDEs, which include changes to PDE parameters ($\gamma$ and $\delta$) and the source term / IC $q$; while \textbf{S2} shows effective extrapolation of solution to a longer time domain, i.e., $t\in[0,4]$. For both scenarios, the Baldwinian-PINNs learn the solution accurately in milli-seconds. The average MSE given by a successfully evolved Baldwinian-PINN over all test tasks ($n=87$) for \textbf{S1} and \textbf{S2} are 1.06$\mathrm{e}{-5}$ {\tiny$\pm$1.60$\mathrm{e}{-5}$} and 1.72$\mathrm{e}{-5}$ {\tiny$\pm$3.02$\mathrm{e}{-5}$}, respectively. Illustrative results are in Figure~\ref{fig:linear}.

Interestingly, the Baldwinian-PINNs maintain good accuracy on tasks from \textbf{S2}, whereby the evolved Baldwinian-PINN model first predicts $u(x,t)$ for $t\in[0,2]$, before using the solution $u(x,t=2)$ as new IC for $t\in[2,4]$. This is achievable because Baldwinian neuroevolution does not require parameterization for the tasks, and Baldwinian-PINNs can generalize to new ICs, BCs, and PDE source terms. This is tricky for existing meta-PINN methods as they require interpolation across a potentially infinitely large distribution of tasks (e.g. possible BCs or ICs), in contrast to the lifetime learning encapsulated in the Baldwinian paradigm. Additional results illustrating the ability of Baldwinian-PINNs to perform well despite variations in the lifetime learning task objectives (e.g., solving for different time durations) are in Suppl.~\ref{subsec:additional_linearpde_time}, further emphasizing the merits of Baldwinian neuroevolution for physics. 

\subsubsection{Additional linear ODE/PDE problems}\label{subsec:otherlinearpde}

The Baldwinian neuroevolution of physics is further demonstrated on 3 linear problems (Problems 3-5), namely 1D Poisson's equation, 2D Poisson's equation, and Helmholtz equation (see Suppl.~\ref{subsec:otherlinearpde2}). Suppl. Table~\ref{tab:linearbenchmark} enumerates the accuracy advantages of Baldwinian-PINN relative to results reported by recent meta-learning PINN works~\cite{liu2022novel, cho2024hypernetwork}. Suppl.~\ref{subsec:otherlinearpde2} also presents additional visualization results from Baldwinian-PINN, other instantiations of the proposed Baldwinian neuroevolution framework (such as natural evolution strategies (NES) variant~\cite{nomura2022fast}), and empirical study of Baldwinian-PINN performance when training-task labels are noisy or scarce, demonstrating the Baldwinian-PINNs' versatility and generalizability. 

\subsection{Learning to solve and generalize nonlinear ODE/PDEs}\label{subsec:nonlinearproblems}


\begin{figure*}[h]%
\centering
\includegraphics[width=1.0\textwidth]{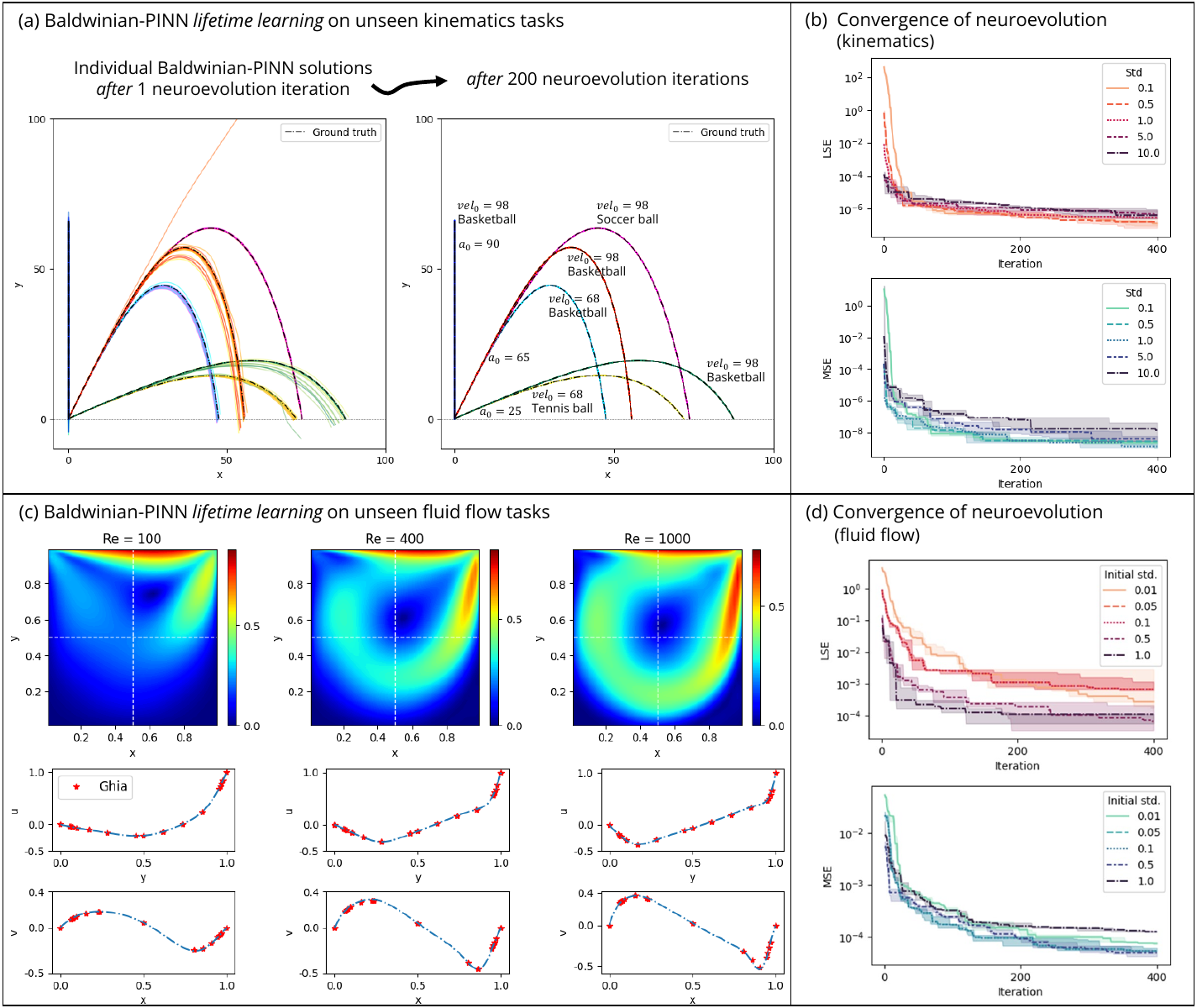}
\vspace{-0.6cm}
\caption{(a) Solution of 20 individual Baldwinian-PINN models sampled from the CMA-ES search distribution (initial std. = 0.5), for unseen kinematics tasks. (c) Solution of Baldwinian-PINN model sampled from the center of NES search distribution after 400 iterations (initial std. = 0.5), for unseen fluid flow (N-S equations) tasks. The velocity magnitude predictions are visually indistinguishable from the ground truth, even as Reynolds number (Re) increases. Their mid‑section $u$ and $v$-velocity profiles (dashed lines) show excellent agreement with the classic benchmark results (marked points) from Ghia~\cite{ghia1982high}. Baldwinian neuroevolution demonstrates effective LSE and MSE convergence on both (b) kinematics and (d) fluid flow problem, for different CMA-ES initial std. values. The bold lines indicate their median convergence path from 5 individual runs, and the shaded areas indicate their interquartile ranges.}\label{fig:nonlinear}
\end{figure*}


\subsubsection{Kinematics}\label{subsec:nonlinearode}

Extending beyond linear ODE/PDEs, Baldwinian-PINNs are applied to nonlinear kinematics equations. Assuming a ball is thrown at specific launch angle $a_0$ and initial velocity $vel_0$, the following 2D kinematics equations describe the ball's motion under the influence of gravity $g$ and air resistance $R$:
\begin{subequations} \label{eq:projectile}
    \begin{multline} 
        \text{(Problem 6)} \quad \frac{d^2x}{dt^2} + R \frac{dx}{dt} = 0, \quad \quad t \in (0, T] 
        \label{eq:projectile_1}
    \end{multline}
    \begin{multline} 
       ~~~~~~~~~~~~~~~~~ \frac{d^2y}{dt^2} + R \frac{dy}{dt} = -g, \quad \quad t \in (0, T]
        \label{eq:projectile_2}
    \end{multline}
\end{subequations}
subject to ICs $x(t=0)=0, \frac{dx}{dt}(t=0)=vel_0\times\mathrm{cos}(\frac{a_0 \pi}{180})$; $y(t=0)=0, \frac{dy}{dt}(t=0)=vel_0\times\mathrm{sin}(\frac{a_0 \pi}{180})$. The air resistance $R=\frac{1}{2}\frac{\rho C_d A}{m}V$ is related to air density $\rho$, object properties (drag coefficient $C_d$, cross-sectional area $A$, and mass $m$), and object velocity $V = \sqrt{(dx\mathrm{/}dt)^2+(dy\mathrm{/}dt)^2}$, hence the equations are nonlinear with respect to $x$ and $y$. The 150 training tasks comprise different launch angles $a_0\in[15,85]$, initial velocity $vel_0\in[10,110]$, and object properties $C_d\in[0.2,0.7]$, $A\in[0.00145,0.045]$, $m\in[0.046,0.6]$ as may be representative of different projectiles (e.g., baseball, basketball). $g$ and $\rho$ are assumed to be $9.8$ and $1.3$ respectively.

The Baldwinian-PINN learns the horizontal and vertical position $x(t)$ and $y(t)$ of the object via iterative least-squares computation (see Section~\ref{subsec:frpinn}) with a fixed number of nonlinear iterations $N = 15$. 100 test tasks encompassing different $a_0\in[5,90]$, $vel_0\in[8,98]$, and $C_d$ are constructed to assess generalizability. Results presented in Figure~\ref{fig:nonlinear}a-b show that Baldwinian-PINNs learn very accurate solutions (MSE $=$ 2.2$\mathrm{e}{-8}$ {\tiny$\pm$1.3$\mathrm{e}{-7}$}) in milli-seconds.

\subsubsection{Set of nonlinear PDEs}\label{subsec:nonlinearpde}

The Baldwinian neuroevolution of physics is further demonstrated on 5 nonlinear PDE problems (Problems 7-11) such as the Burger's, nonlinear Allen-Cahn and nonlinear reaction-diffusion equations, as per recent meta-learning PINN study \cite{penwarden2023metalearning} and described in Suppl.~\ref{subsec:nonlinearpde2}. That study compares several meta-learning PINNs based on weight interpolation methods and MAML.


\begin{figure*}[h]%
\centering
\includegraphics[width=1.0\textwidth]{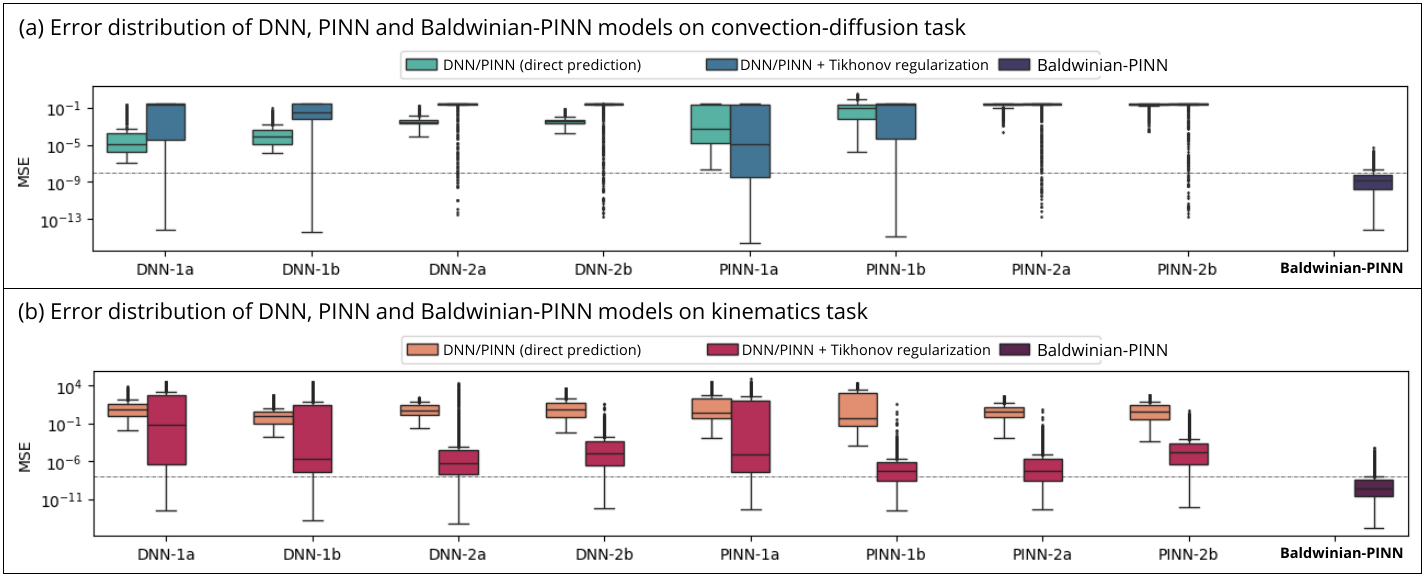}
\vspace{-0.6cm}
\caption{The generalization performance of the DNN, PINN, and Baldwinian-PINN models are compared, across (a) 110 convection-diffusion and (b) 100 kinematics test tasks. Model-\textbf{1a/2a}: deep/shallow architecture with \textit{tanh} activations (network weights initialized by \textit{Xavier} method); Model-\textbf{1b/2b}: deep/shallow architecture with \textit{sin} activations (weights initialized by \textit{He} method). The MSE results from each of the DNN and PINN models are pooled from 3 initial learning rate configurations $\times$ 5 individual runs. The MSE results from Baldwinian-PINN model are pooled from 5 initial std. values $\times$ 5 individual runs.}\label{fig:ablation}
\end{figure*}


A key difference between this study and meta-learning of PINNs in \cite{penwarden2023metalearning} is the availability of data for training tasks. On the test tasks, the evolved Baldwinian-PINNs can achieve $\backsim 1$ order of magnitude lower relative error compared to \cite{penwarden2023metalearning}. Crucially, the computation time for Baldwinian-PINN for new predictions is at most two seconds whereas existing meta-learning approaches may take more than 500 seconds (2 orders of magnitude acceleration). For example, when solving the diffusion-reaction equation, a 70x improvement in accuracy was achieved relative to other state-of-the-art approaches, while taking 700x less computational time. Complete quantitative results are summarized in Suppl. Table~\ref{tab:benchmark}.

The versatility of Baldwinian neuroevolution is also key here as we can accelerate the Baldwinian-PINN's lifetime learning by using a much coarser discretization and smaller number of nonlinear iterations during training while still learning a good solution on test tasks with finer (e.g., 16$\times$) discretization and more (e.g., 2$\times$) nonlinear iterations. 

\subsubsection{Navier-Stokes (N-S) equations for fluid flow}\label{subsec:nonlinearns}

We explore a more challenging physics scenario in computational fluid dynamics, the lid-driven cavity flow problem. The fluid flow inside a 2D unit square, $x \in [0,1]$, $y \in [0,1]$, is driven by the top lid velocity ($u_{lid} = 1$), and governed by the steady-state incompressible N-S equations for dependent variables velocity $u,v$ and pressure $p$:
\begin{subequations}
    \label{eq:NS}
    \begin{align}
        \label{eq:continuity}
        \nabla\cdot \vec{u} &= 0 \\
        \label{eq:momentum}
        (\vec{u}\cdot\nabla)u &= \frac{1}{Re}\nabla^2 \vec{u} - \nabla p
    \end{align}
\end{subequations}
Complex flow structures emerge at moderate to high Reynolds numbers, making this a challenging benchmark for PINNs, which may require hours of training with additional techniques such as transfer or curriculum learning to achieve convergence.

The Baldwinian neuroevolution of physics is performed on 7 training tasks $Re\in[1,800]$. After 400 neuroevolution iterations, the evolved Baldwinian-PINNs can generalize to new solutions $Re\in[5,1000]$ with an average MSE of 2.1$\mathrm{e}{-4}$ {\tiny$\pm$2.4$\mathrm{e}{-4}$} ($n=18$ tasks $\times$ 5 individual runs). Results presented in Figure~\ref{fig:nonlinear}c-d and Suppl.~\ref{subsec:additional_nsflow} Figure~\ref{fig:additional_nsflow} demonstrate excellent agreement between Baldwinian-PINN predictions and the numerical benchmark from Ghia \textit{et al.}~\cite{ghia1982high}. Baldwinian-PINNs learn accurate solutions to the N–S equations in under one second, highlighting the potential to address difficult physics problems characterized by strong nonlinearity and strong coupling among multiple variables.

\subsection{Analysis on effectiveness of Baldwinian neuroevolution}\label{subsec:analysis}

We further investigate the advantages of Baldwinian neuroevolution via an ablation study using the \textit{convection-diffusion} and \textit{kinematics} examples.

Briefly, we explore deep and shallow MLP architectures for baseline SGD-trained DNN and PINN models for comparison with the Baldwinian-PINNs: \textbf{1.} the deep architecture consists of similar total number of network weights as the corresponding Baldwinian-PINN models, but distributed across multiple nonlinear hidden layers with smaller number of nodes; \textbf{2.} the shallow architecture has single nonlinear hidden layer and same number of nodes as the corresponding Baldwinian-PINN model but this necessitates more total network weights because of the additional input variables.

Each architecture also consists of 2 variants: \textbf{a.} \textit{tanh} activation for the nonlinear hidden layers whereby the network weights are initialized by \textit{Xavier} method; \textbf{b.} \textit{sin} activation for the nonlinear hidden layers whereby the network weights are initialized by \textit{He} method.

This notation is maintained when referencing the respective model performance in Figure~\ref{fig:ablation}. For example, the models labelled as DNN-1a and PINN-2b refer to the corresponding baseline DNN model with a deep architecture and \textit{tanh} activation layer and baseline PINN model with a shallow architecture (single hidden layer) and \textit{sin} activation layer respectively. Additional model descriptions are in Suppl.~\ref{subsec:baseline}. 

\subsubsection{Direct prediction across tasks with parametric DNNs / PINNs}

As a baseline, DNN and PINN models are trained with SGD (ADAM) based on data-driven loss and PINN (data and physics) loss respectively, and applied to predictions for unseen tasks based on interpolation. We explore different MLP configurations as per Suppl. Table~\ref{tab:baseline_model} in order to more fairly compare the best performance across different models.

Critically, when making a direct prediction for a new task, the respective baseline DNN and PINN models do not explicitly incorporate (and guarantee compliance with) the known physics prior. Although the parametric PINN can learn a more physically-consistent prediction from the physics loss during training, compliance is not guaranteed for a new task. 

In addition, the new tasks must follow an \textit{a priori} determined input parameterization for interpolation. For example, the DNN and PINN models for convection-diffusion need to be \textit{a priori} set-up with $(x, \alpha)$ as inputs to enable predictions across different $\alpha$'s. In contrast, the Baldwinian-PINN is parameterization-agnostic, and does not require any input parameterization for predictions of new tasks.

The \textit{direct prediction} performance of DNNs and PINNs with different model configurations are summarized in Figure~\ref{fig:ablation} and Suppl. Table~\ref{tab:baseline_results}. These baseline DNN and PINN predictions have high generalization errors (several orders of magnitude higher than Baldwinian-PINN). As explained above, PINNs do not necessarily outperform DNNs for direct prediction on test tasks (e.g., MSE for convection-diffusion for different model configurations range from $1.7\mathrm{e}{-4}$ to $1.2\mathrm{e}{-2}$ for DNN, and from $1.9\mathrm{e}{-3}$ to $3.4\mathrm{e}{-1}$ for PINN), suggesting that PINN training by itself does not always guarantee generalization in the absence of additional physics-informed fine-tuning. 

\subsubsection{Prediction across tasks with parametric DNNs / PINNs and physics-informed fine-tuning}

In order to assess the utility of a physics-informed fine-tuning step, we optimize the final layer weights of the SGD-trained baseline DNNs and PINNs with the Tikhonov regularization at test time. This fine-tuning procedure is almost identical to the Baldwinian-PINN's lifetime learning. 

The key difference is that the pre-final nonlinear hidden layers weight distribution and learning hyperparameter $\lambda$ in Baldwinian-PINNs are jointly learned by evolution to share among tasks. In contrast, nonlinear hidden layers' weights for the DNNs and PINNs are SGD-learned from train tasks. As the Tikhonov regularization solution is highly dependent on $\lambda$, we need to perform a grid search across $\lambda = \{1\mathrm{e}{-2}, 1\mathrm{e}{-4}, 1\mathrm{e}{-6}, 1\mathrm{e}{-8}, 1\mathrm{e}{-10}, 1\mathrm{e}{-12}, 0\}$ to determine the best $\lambda$ during test time and present the solution associated with the lowest LSE. The results \textit{after Tikhonov regularization} are compared in Figure~\ref{fig:ablation} and Suppl. Table~\ref{tab:baseline_results}.

Results show that physics-informed fine-tuning via Tikhonov regularization can improve the accuracy of both data-trained DNN and physics-trained PINN models at test time. However, their results are not as generalizable (i.e., the improvements are isolated to certain model and task instances) as the Baldwinian-PINN (which has been optimized for generating accurate solution over task distribution). 

For example, we observe a significant improvement for the nonlinear kinematics problem (i.e., the best MSE for DNN and PINN improves from $3.0$ to $4.4\mathrm{e}{-5}$ and $1.3$ to $7.9\mathrm{e}{-6}$ respectively after Tikhonov regularization), potentially because the training distribution is sparser relative to the larger variation in output patterns (see Suppl.~\ref{subsec:additional_ablation}). 

In contrast, fine-tuning via Tikhonov regularization doesn't produce better results than direct prediction for the convection-diffusion problem. The MSE with and without Tikhonov regularization are $2.3\mathrm{e}{-4}$ and $2.8\mathrm{e}{-2}$, and $1.9\mathrm{e}{-3}$ and $7.2\mathrm{e}{-3}$ for the best-configured DNN and PINN respectively. Tikhonov regularization cannot jointly minimize both PDE and IC/BC errors to sufficiently small levels, leading to inferior performance than direct prediction for some tasks. 

While the best DNN model (after Tikhonov regularization) can be more accurate than the corresponding Baldwinian-PINN on some convection-diffusion tasks at lower $\alpha$, solution quality deteriorates quickly at larger $\alpha$. Interestingly, PINNs with the same configuration (after Tikhonov regularization) show the opposite trend, highlighting the challenge of obtaining a generalizable model for entire task distributions through SGD, in contrast to Baldwinian neuroevolution.

Overall, these results suggest that an additional step of physics-informed learning (i.e. Tikhonov regularization) on any new task can be beneficial. However, SGD-trained PINN and DNN models learn pre-final nonlinear hidden layers that are highly variable in their suitability for the physics-informed Tikhonov regularization-based fine-tuning across new tasks, in contrast to the proposed Baldwinian-PINNs. Hence, these results highlight the advantages of Baldwinian-PINN framework as a complete solution to the challenges encountered in generalizing with physics-informed learning.  

\section{Conclusion}\label{sec:discussion}

In this paper, we study the Baldwin effect as a novel means to advancing the generalizability of PINNs over a family of governing differential equations. The Baldwin effect is instantiated through a two-stage stochastic programming formulation, wherein the first stage evolves the initial layers of a generalizable PINN model, and the second stage trains its final layer to specialize to any new physics task (analogous to lifetime learning). The Baldwinian neuroevolution biases pre-final hidden layers’ weight distribution so that final layer specialization is both fast and accurate, preserving (near) constant-time solves while allowing capacity to scale. Although the pre-final hidden layers are not fine-tuned per task, the Baldwinian-PINN’s capacity can be effectively scaled---by increasing width and/or depth---to handle more complex physics (e.g., fluid flow) while preserving fast test-time learning.

Our method is demonstrated to be broadly applicable to the learning of different linear and nonlinear ODE/PDEs encompassing diverse physical phenomena such as convection-diffusion, particle kinematics, heat and mass transfer, and fluid flow. Relative to recent meta-learned PINNs, Baldwinian-PINNs can accelerate the physics-aware predictions by several orders of magnitude, while improving the prediction accuracy by up to one order. A Baldwinian-PINN is thus in the image of a precocial species with accelerated learning ability at birth.
 
The lifetime learning encapsulated in the Baldwinian paradigm does not require \textit{a priori} parameterization for the task scenarios, permitting both flexible generalization to new ICs, BCs, and PDE source terms and variations in the lifetime learning task objectives (e.g., different domain and/or sample size). This allows Baldwinian-PINNs to be useful in applications when large number of evaluations with \textit{a priori} unknown input conditions are sought, e.g. generative design or what-if analysis. In addition, results in Section~\ref{subsec:linearpde} suggest that Baldwinian-PINNs could be suited to the continual modelling of dynamical systems through their versatility in handling ICs and ability to rapidly model and stitch time windows together with minimal error~\cite{meng2020ppinn}.

In the context of recent interest in \emph{foundation models for scientific machine learning}~\cite{subramanian2023towards}, specifically through the use of neural operators or neural PDE solvers, our experiments showing accurate and fast generalization across families of linear and nonlinear ODE/PDEs suggest that Baldwinian learning can be an alternate route to such flexible and generalizable machine intelligence models. It will be interesting to test the limits to which Baldwinian-PINNs can learn across broad classes of physics phenomena and/or differential operators in future work. Hidden-layer fine-tuning adds flexibility but forfeits the closed-form update and re-introduces slow, nonconvex optimization. For large task shifts (out-of-distribution), a light hidden-layer adaptation (e.g., a few gradient-descent steps on a low-rank adapter) are a potential practical extension that trades some speed for additional flexibility.

Lastly, while the experiments in this work focus on optimizing the center and spread of the probability distributions that sample the initial weights in the neural network layers, this can be easily extended to incorporate other state-of-the-art neural architecture search approaches, which directly optimize the graph structure of the node connections. Our experiments indicate that significant improvement relative to other recent meta-learning PINN works can already be observed even when we only optimize the center and spread of the weight sampling distributions under the Baldwinian-PINN framework. The proposed Baldwinian-PINN framework can be seamlessly extended in future work, e.g. via integration with other state-of-the-art neural architecture search algorithms.

\section*{Acknowledgment}
This research was in part supported by the National Research Foundation, Singapore through the AI Singapore Programme, under the project ``AI-based urban cooling technology development" (Award No. AISG3-TC-2024-014-SGKR). Abhishek Gupta was supported by the Ramanujan Fellowship from the Anusandhan National Research Foundation, Government of India (Grant No. RJF/2022/000115).






\bibliographystyle{IEEEtran}
\bibliography{IEEEabrv,Main}

\clearpage


\begin{sidewaystable*}

\caption{Summary of Baldwinian neuroevolution and Baldwinian-PINN’s lifetime learning configurations, and generalization performance on PINN problems in experimental study.} \label{tab:summary}
\begin{tabular*}{\textheight}{@{\extracolsep\fill}l>{\raggedright\arraybackslash}p{3cm}p{1cm}>{\raggedright\arraybackslash}p{1cm}p{1cm}p{1cm}p{1cm}p{1cm}p{1cm}p{1cm}>{\raggedright\arraybackslash}p{2.2cm}p{1cm}p{1cm}p{0.9cm}p{1.2cm}}
\toprule%
& \multicolumn{2}{@{}c@{}}{\textbf{Task distribution}} & \multicolumn{6}{@{}c@{}}{\textbf{Baldwinian neuroevolution w/ CMA-ES or NES}} & \multicolumn{3}{@{}c@{}}{\textbf{Baldwinian-PINN lifetime learning}} & \multicolumn{3}{@{}c@{}}{\textbf{Predictive performance}} \\ \cmidrule(lr){2-3} \cmidrule(lr){4-9} \cmidrule(lr){10-12} \cmidrule(lr){13-15}%
& Problem & No. train tasks & Batch size for task, $n_{task}$ & Population size, $n_{pop}$ & Max. iteration & Initial std. & Search dimension for $(\theta,\lambda)$ & Time\footnotemark[1] (s) & dim($\vb*{w}$) & Sample size $(n_{pde}, n_{ic}, n_{bc})$ & nonlinear iterations, $N$ & No. test tasks & Time\footnotemark[2] (s) & MSE\footnotemark[3] \\
\midrule
1 & Convection-diffusion $x \rightarrow u$ & 20 & 10 & 20 & 200 & 1 & 25 & 250 & 900 & 1001, -, 2 & - & 110 & 0.005 & 5.8e-9 {\tiny$\pm$ 9.9e-9} \\[0.2cm]
2 & Family of linear PDEs $(x, t) \rightarrow u$ & 108 & 15 & 20 & 500 & 5 & 37 & 10830 & 1200 & 5151, 101, 50 {\tiny(10201, 101, 100)\footnotemark[4]} & - & 87 & 0.122 & 7.1e-4 {\tiny$\pm$ 3.1e-3} \\[0.2cm]
3 & 1D Poisson's equation $x \rightarrow u$ & 60 & 10 & 20 & 100 & 1 & 25 & 120 & 900 & 1001, -, 2 & - & 60 & 0.019 & 1.0e-9 {\tiny$\pm$ 5.8e-9} \\[0.2cm]
4 & 2D Poisson's equation $(x, y) \rightarrow u$ & 100 & 20 & 20 & 100 & 1 & 37 & 270 & 900 & 1089, -, 128 {\tiny(16641, -, 512)\footnotemark[5]} & - & 100 & 0.132 & 8.9e-12 {\tiny$\pm$ 1.0e-11} \\[0.2cm]
5 & Helmholtz equation $(x, y) \rightarrow u$ & 20 & 10 & 20 & 400 & 5 & 38 & 520 & 900 & 1024, -, 124 {\tiny(16384, -, 508)\footnotemark[5]} & - & 60 & 0.131 & 1.6e-5 {\tiny$\pm$ 5.1e-5} \\[0.2cm]
6 & Kinematics $t \rightarrow (x, y) \quad\quad$ & 150 & 30 & 20 & 400 & 0.5 & 25 & 1600 & 1800 & 101, 1, - & 15 & 100 & 0.039 & 2.2e-8 {\tiny$\pm$ 1.3e-7}   \\[0.2cm]
7 & Burgers’ equation $(x, t) \rightarrow u$ & 16 & 5 & 20 & 200 & 1 & 37 & 5240 & 900 & 13107, 257, 100 {\tiny(25957, 257, 200)\footnotemark[5]} & 5 {\tiny(10)\footnotemark[6]} & 32 & 1.96 & 2.3e-7 {\tiny$\pm$ 9.6e-7} \\[0.2cm]
8 & Nonlinear heat $(x, t) \rightarrow u$ & 13 & 5 & 20 & 100 & 5 & 37 & 440 & 900 & 1600, 64, 50 {\tiny(25600, 256, 200)\footnotemark[5]} & 5 & 64  & 0.87 & 1.3e-7 {\tiny$\pm$ 4.1e-7} \\[0.2cm]
9 & Allen-Cahn equation $(x, y) \rightarrow u$ & 16 & 8 & 20 & 100 & 0.5 & 37 & 510 & 900 & 1024, -, 128 {\tiny(16384, - ,512)\footnotemark[5]} & 5 {\tiny(10)\footnotemark[6]} & 32 & 1.19 & 2.0e-7 {\tiny$\pm$ 6.1e-7} \\[0.2cm]
10 & Diffusion-reaction $(x, y) \rightarrow u$ & 22 & 10 & 20 & 100 & 1 & 37 & 640 & 900 & 1024, -, 128 {\tiny(16384, - ,512)\footnotemark[5]} & 5 & 64 & 0.58 & 1.2e-8	{\tiny$\pm$ 8.1e-8} \\[0.2cm]
11 & 6D diffusion-reaction $(x, y) \rightarrow u$ & 17 & 8 & 20 & 100 & 1 & 37 & 510 & 900 & 1024, -, 128 {\tiny(16384, - ,512)\footnotemark[5]} & 5 & 100 & 0.60 & 1.6e-8 {\tiny$\pm$ 2.5e-8} \\[0.1cm]
12 & Fluid flow $(x, y) \rightarrow (u, v, p) \quad\quad$ & 7 & 4 & 20 & 400 & 0.1 & 66 & 15868 & 1152 & 2401, -, 196 & 5 & 18 & 0.903 & 2.1e-4 {\tiny$\pm$ 2.4e-4}   \\[0.1cm]
\end{tabular*}
\footnotetext[1]{Baldwinian neuroevolution time cost, on 2 GPUs (NVIDIA GeForce RTX 3090).}
\footnotetext[2]{Computation time per task, on single GPU (NVIDIA GeForce RTX 3090).}
\footnotetext[3]{MSE results are aggregated from 5 individual runs.}
\footnotetext[4]{Prediction of test task includes projection to a longer (2$\times$) time domain.}
\footnotetext[5]{Prediction of test task on denser sample points.}
\footnotetext[6]{Prediction of test task with more nonlinear iterations.}

\end{sidewaystable*}

\clearpage

\end{document}


\pagenumbering{roman}

\title{Supplementary Material for\\Evolutionary Optimization of Physics-Informed Neural Networks: Advancing Generalizability by the Baldwin Effect}


\author{Jian Cheng Wong, Chin Chun Ooi, Abhishek Gupta, \textit{Senior Member, IEEE}, Pao-Hsiung Chiu, \\ Joshua Shao Zheng Low, My Ha Dao, and Yew-Soon Ong, \textit{Fellow, IEEE}
\thanks{Jian Cheng Wong, Chin Chun Ooi, and Pao-Hsiung Chiu are with the Institute of High Performance Computing, Agency for Science, Technology and Research, Singapore (e-mail: wongj@a-star.edu.sg; ooicc@a-star.edu.sg; chiuph@a-star.edu.sg).}
\thanks{Abhishek Gupta is with the School of Mechanical Sciences, Indian Institute of Technology Goa, India (e-mail: abhishekgupta@iitgoa.ac.in).}
\thanks{Joshua Shao Zheng Low is with the College of Computing and Data Science, Nanyang Technological University, Singapore (e-mail: joshualow188@gmail.com).}
\thanks{My Ha Dao is with the Technology Centre for Offshore and Marine, Singapore (e-mail: Dao_My_Ha@tcoms.sg).}
\thanks{Yew-Soon Ong is with the Agency for Science, Technology and Research, Singapore, and is also with the College of Computing and Data Science, Nanyang Technological University, Singapore (e-mail: Ong_Yew_Soon@a-star.edu.sg).}}

\maketitle
\IEEEpeerreviewmaketitle

\renewcommand{\thesection}{S.\Roman{section}} 
\renewcommand{\thesubsection}{\thesection.\Alph{subsection}}



\renewcommand{\theequation}{S\arabic{equation}}

\makeatletter
\renewcommand\thetable{S\@arabic\c@table}
\renewcommand \thefigure{S\@arabic\c@figure}
\makeatother


\section{Data Generation}\label{subsec:data}

In this study, the majority of the problems have corresponding analytical solutions as described in the problem setups. In addition, the ground truth for the PDE family (Problem 2) and Burgers’ equation (Problem 7) is obtained by a high-resolution finite volume scheme. To alleviate convection instability, the dispersion-relation-preserving (DRP) finite volume scheme with a universal limiter has been utilized \cite{Chiu2018,Leonard1991}. Other spatial derivative terms are discretized by central difference. For the temporal term, the second-order TVD Runge Kutta scheme \cite{Gottlieb98} is employed. For the incompressible Navier-Stokes (N-S) equations (Problem 12), the dispersion-relation-preserving (DRP) finite volume scheme, together with improved divergence-free-condition compensated framework~\cite{Chiu2018}, is employed to produce high-fidelity solutions as well as ensure the velocity-pressure coupling.

For the PDE family equation, the spatial resolution, $\Delta x$, is $1/400$, while temporal resolution, $\Delta t$, is $5\mathrm{e}{-5}\Delta x$; For the Burgers’ equation, the spatial resolution, $\Delta x$, is $1/512$, while temporal resolution $\Delta t$, is $1\mathrm{e}{-2}\Delta x$; for the incompressible N-S equations, the spatial resolution, $\Delta x$ and $\Delta y$, are $1/200$.

\section{Lagging of Coefficient Method for Nonlinear PDEs}\label{subsec:lagging}

For nonlinear PDEs, iterative methods can be used for arriving at optimized Baldwinian-PINNs. In this work, we use a \textit{lagging of coefficient approach} which is common in numerical methods \cite{anderson2020computational}. Briefly, we approximately linearize the nonlinear term(s) in Eq. \ref{eq:pde_ibc_eqn} by substituting the output $u(x,t) = \sum_j w_j f_j(x,t;\vb*{\tilde{w}})$ obtained from previous step, and iteratively solve Eq. \ref{eq:pseudoinverse} to update $w_j$'s for a fixed number of steps, $N$, or until a convergence criterion is reached.

The nonlinear equation $(\frac{d^2u}{dx^2} + \frac{d^2u}{dy^2}) + u(1-u^2) = f$ is used to demonstrate the \textit{lagging of coefficient} method for computing the $w_j$’s in Baldwinian-PINNs. We approximately linearize the nonlinear term $u(1-u^2)$ as $u(1 - \check{u}^2)$ with $\check{u}=0$ being the initial guess solution at first iteration. The $(i\text{-th PDE sample},j\text{-th neuron})$ entry of the least squares matrix $\vb{A}$ in Eq. \ref{eq:leastsquares} (Section~\ref{subsec:frpinn}) now becomes:
\begin{equation}
    \begin{split}
        \mathcal{N}_\theta[f_j(x_{i}^{pde},y_{i}^{pde};\vb*{\tilde{w}}_j)] = &\ \frac{d^2f_j(x_{i}^{pde},y_{i}^{pde};\vb*{\tilde{w}}_j)}{dx^2} \\
        & + \frac{d^2f_j(x_{i}^{pde},y_{i}^{pde};\vb*{\tilde{w}}_j)}{dy^2} \\
        & + f_j(x_{i}^{pde},y_{i}^{pde};\vb*{\tilde{w}}_j)(1 - \check{u}^2)
    \end{split}
    \label{eq:lagging_1}
\end{equation}

Hence, the Tikhonov regularization solution $\vb*{w}$ can be obtained from the linearized version of the equation. The solution obtained from past iteration is used to compute $\check{u}(x,y) = \sum_j w_j f_j(x,y;\vb*{\tilde{w}}_j)$ for the next iteration, until the solution $\vb*{w}$ reaches a specified convergence criterion or reaches a pre-determined number of iterations, $N$.

\section{Additional Results for Family of Linear PDE Problem}


\begin{figure*}[h]%
\centering
\includegraphics[width=1.0\textwidth]{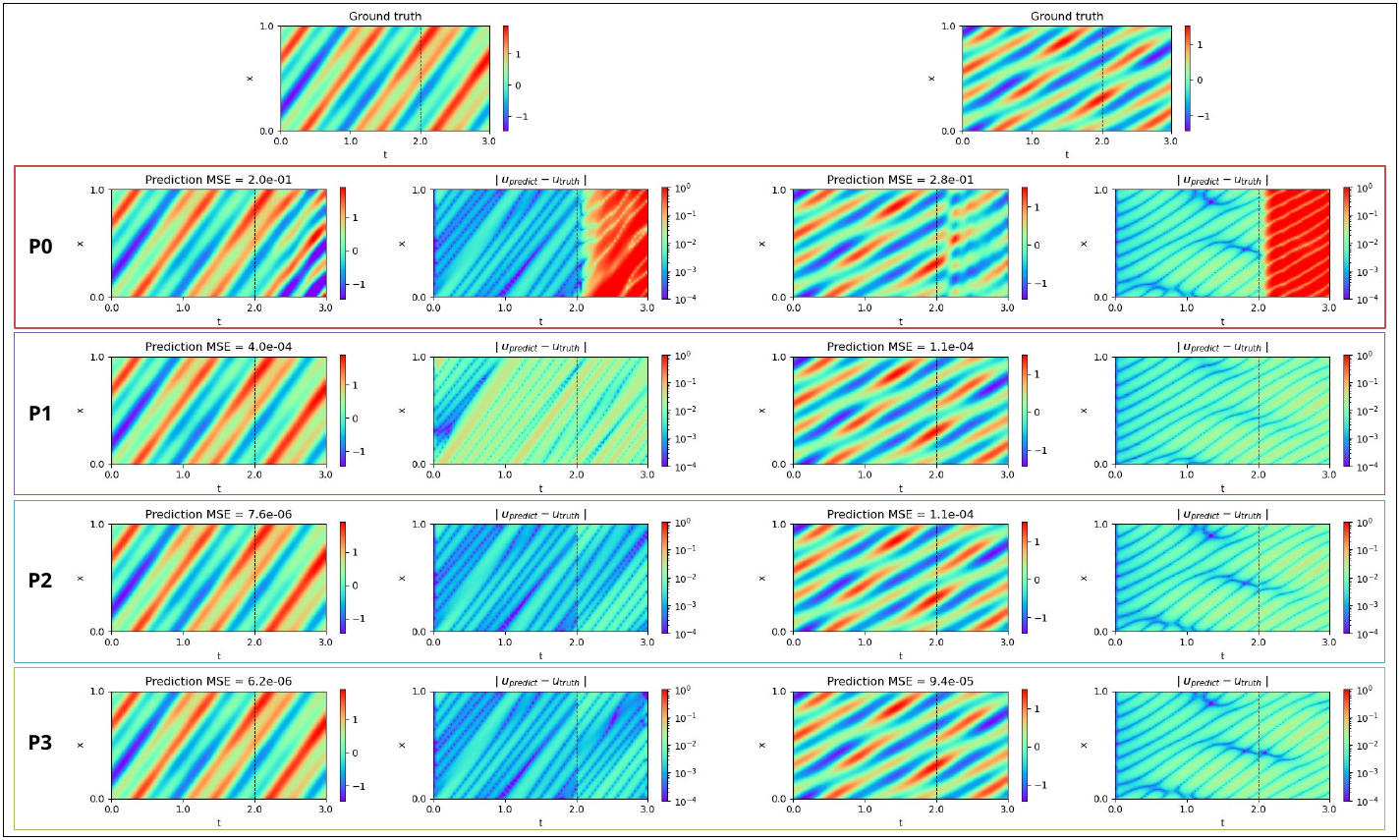}
\vspace{-0.4cm}
\caption{Solution for unseen family of linear PDE tasks on $t\in[0,3]$ obtained by best evolved Baldwinian-PINN (sampled from the center of CMA-ES search distribution after 500 iterations with initial std. = 5) using different ways \textbf{P0-P3}.}\label{fig:additional_linear_time}
\end{figure*}



\begin{figure*}[h]%
\centering
\includegraphics[width=1.0\textwidth]{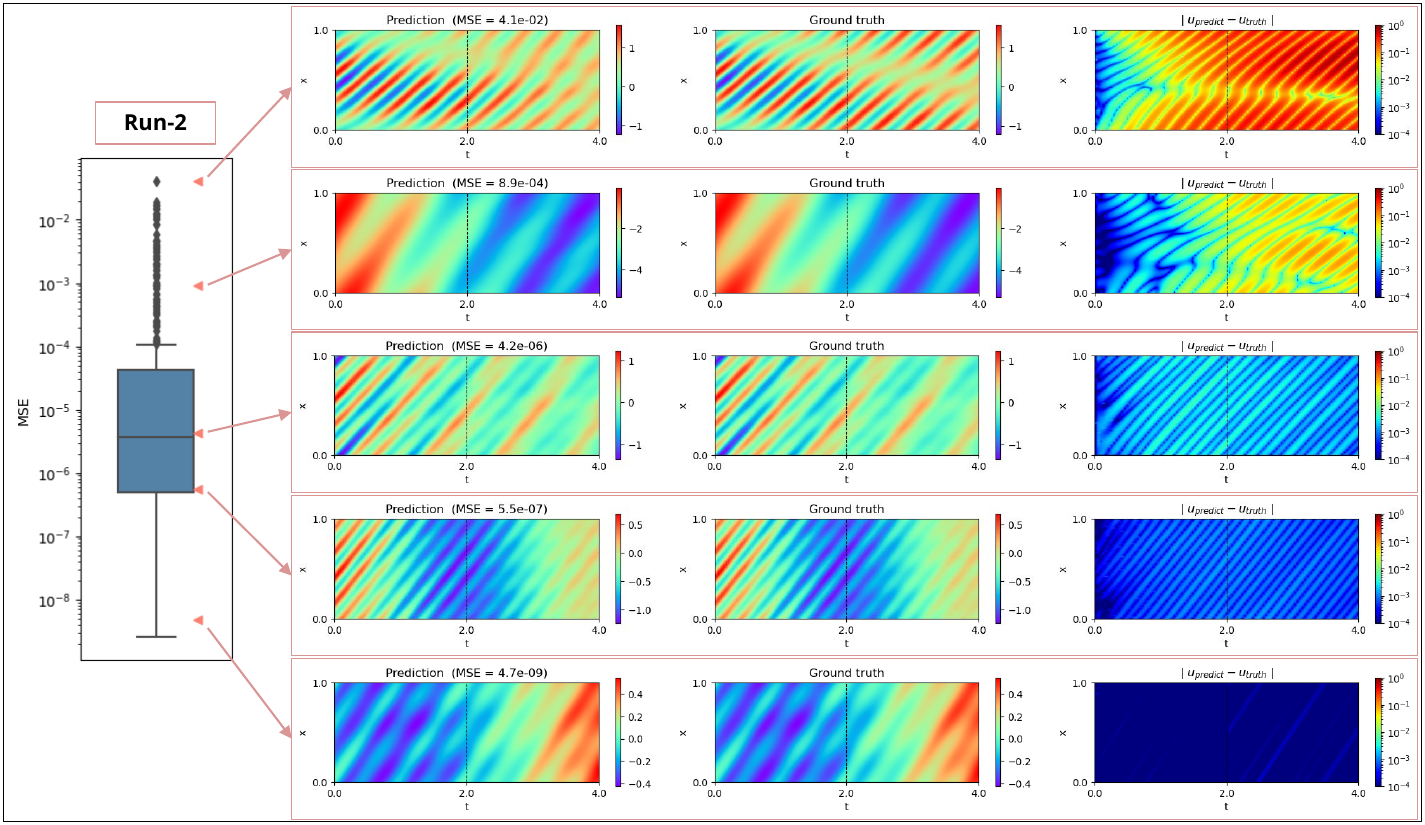}
\vspace{-0.4cm}
\caption{Baldwinian-PINN's solution (\textbf{run no. 2}) vs. ground truth on 5 selected family of linear PDE tasks, and the position of their accuracy along the MSE spectrum (pooled from $n=87$ test tasks $\times$ 5 individual runs with initial std. = 5).}\label{fig:additional_linear_run2}
\end{figure*}

\begin{figure*}[h]%
\centering
\includegraphics[width=1.0\textwidth]{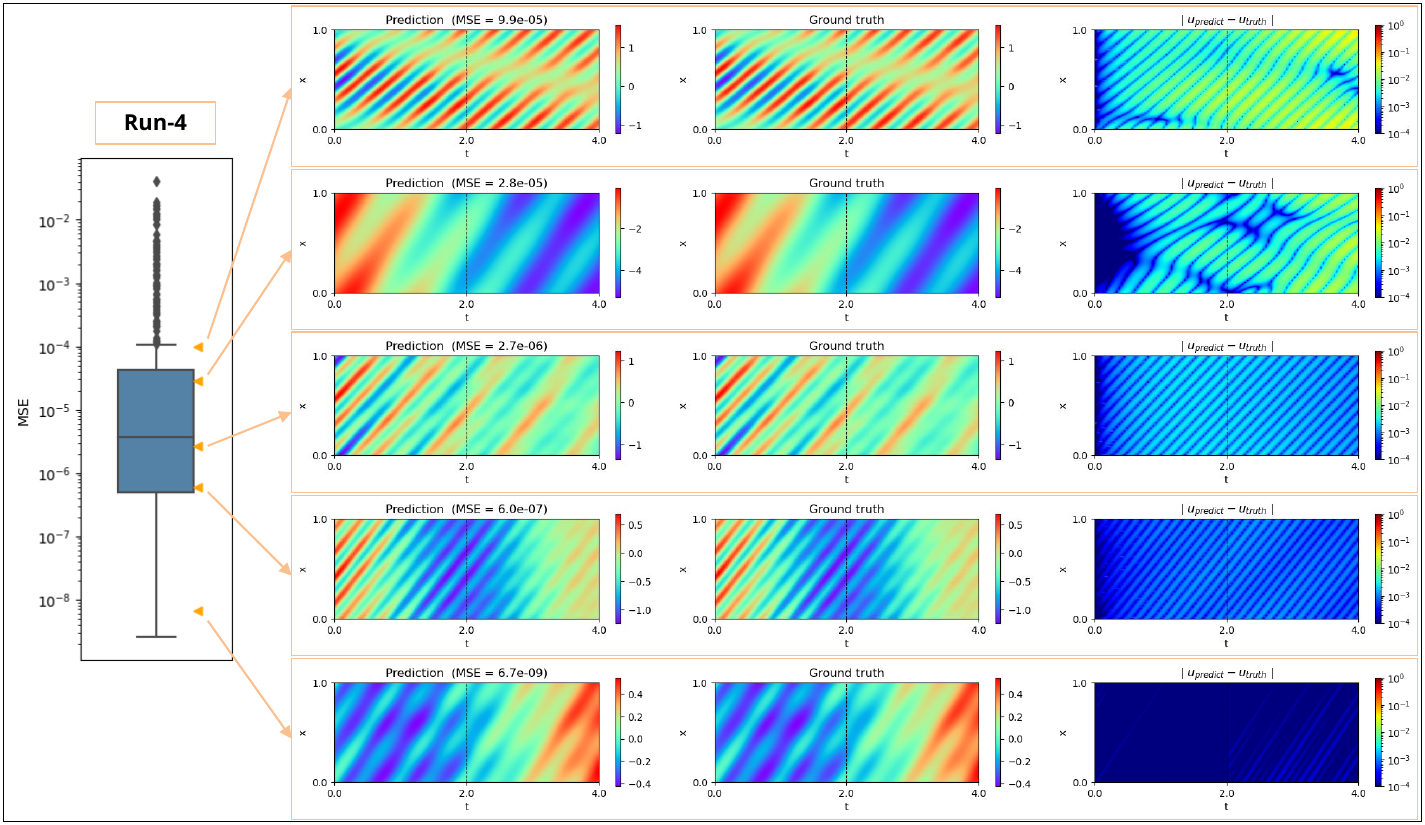}
\vspace{-0.4cm}
\caption{Baldwinian-PINN's solution (\textbf{run no. 4}) vs. ground truth on 5 selected family of linear PDE tasks, and the position of their accuracy along the MSE spectrum (pooled from $n=87$ test tasks $\times$ 5 individual runs with initial std. = 5).}\label{fig:additional_linear_run4}
\end{figure*}


\subsection{Prediction on new time interval}\label{subsec:additional_linearpde_time}

The experimental results in Section~\ref{subsec:linearpde} show that Baldwinian-PINNs trained on a set of linear PDE tasks for $t\in[0,2]$, are capable of learning time-dependent solution $u(x,t)$ on a set of test tasks for unseen PDEs and ICs, as well as for a longer (2 $\times$) time domain by performing the learning twice, i.e., for $t\in[0,2]$ and then using the learned solution $u(x,t=2)$ as new IC for $t\in[2,4]$. Recall that the average MSE given by the best evolved Baldwinian-PINN model (sampled from the center of CMA-ES search distribution from the best run) over all test tasks for $t\in[0,2]$ and $t\in[0,4]$ are 1.06$\mathrm{e}{-5}$ {\tiny$\pm$1.60$\mathrm{e}{-5}$} and 1.72$\mathrm{e}{-5}$ {\tiny$\pm$3.02$\mathrm{e}{-5}$}, respectively.

We further demonstrate the versatility of Baldwinian-PINNs by changing the time domain of interest in the test tasks to $t\in[0,3]$. From the best evolved Baldwinian-PINN model, we can obtain the prediction on new tasks in the following ways: \textbf{P0} the solution for the original time domain $t\in[0,2]$ via physics-based lifetime learning, and the extended time $t\in[2,3]$ based on neural network interpolation; \textbf{P1} the solution for original and extended time domain $t\in[0,3]$ altogether via physics-based lifetime learning; \textbf{P2} physics-based learning of the solution for the first 2s time $t\in[0,2]$, before using the learned solution $u(x,t=2)$ as new IC for next 2s time $t\in[2,4]$; \textbf{P3} physics-based learning of the solution for the first 2s time $t\in[0,2]$, before using the learned solution $u(x,t=2)$ as new IC for next time domain $t\in[2,3]$.

Figure~\ref{fig:additional_linear_time} shows the Baldwinian-PINN's solution obtained using different ways \textbf{P0-P3} on selected test tasks. Their MSE results over all test tasks for $t\in[0,3]$ are 5.35$\mathrm{e}{-2}$ {\tiny$\pm$6.08$\mathrm{e}{-2}$} (\textbf{P0}), 4.24$\mathrm{e}{-5}$ {\tiny$\pm$1.04$\mathrm{e}{-5}$} (\textbf{P1}), 1.47$\mathrm{e}{-5}$ {\tiny$\pm$2.32$\mathrm{e}{-5}$} (\textbf{P2}), and 1.43$\mathrm{e}{-5}$ {\tiny$\pm$2.22$\mathrm{e}{-5}$} (\textbf{P3}), respectively. The performance of \textbf{P0} is significantly worse than \textbf{P1-P4}, as expected, because of the physics-agnostic extrapolation. The Baldwinian-PINNs can flexibly learn the solutions on extended time domain in a single Tikhonov regularization solve (\textbf{P1}), although the most accurate solutions are given by \textbf{P2} and \textbf{P3}.

\subsection{Performance variation across runs}

The Baldwinian neuroevolution outcome has the most variation in accuracy across individual runs, on the family of linear PDEs problem relative to the other problems in the experimental studies. The Baldwinian-PINN MSE results over all test tasks for $t\in[0,4]$ obtained by 5 different runs are 5.3$\mathrm{e}{-5}$ {\tiny$\pm$1.8$\mathrm{e}{-4}$}, 1.7$\mathrm{e}{-5}$ {\tiny$\pm$3.0$\mathrm{e}{-5}$}, 2.0$\mathrm{e}{-3}$ {\tiny$\pm$5.6$\mathrm{e}{-3}$}, 1.5$\mathrm{e}{-3}$ {\tiny$\pm$3.5$\mathrm{e}{-3}$}, and 2.8$\mathrm{e}{-5}$ {\tiny$\pm$9.1$\mathrm{e}{-5}$}, respectively. Figures~\ref{fig:additional_linear_run2} and \ref{fig:additional_linear_run4} compare the solutions of 2 Baldwinian-PINNs obtained from separate Baldwinian neuroevolution runs, on the same 5 test tasks selected at different levels of accuracy along the MSE spectrum (pooled from $n=87$ test tasks $\times$ 5 individual runs with initial std. = 5). In all our other experiments, the overall variation across runs remains fairly small.

\clearpage

\begin{sidewaystable*}
\caption{Comparison with baseline meta-learning PINN models on single test task for linear ODE/PDE problems.} \label{tab:linearbenchmark}
\begin{tabular*}{\textheight}{@{\extracolsep\fill}p{0.1cm}>{\raggedright\arraybackslash}p{1.8cm}p{1cm}p{1cm}p{1cm}p{1cm}p{1cm}p{2cm}p{10cm}}
\toprule
{} & \textbf{Problem} & \multicolumn{6}{@{}c@{}}{\textbf{Method / Model}} &  \textbf{Remarks} \\ \cmidrule(lr){3-8}
{} & {} & \multicolumn{2}{@{}c@{}}{NRPINN} & \multicolumn{2}{@{}c@{}}{Hyper-LR-PINN} & \multicolumn{2}{@{}c@{}}{Baldwinian-PINN}  & {} \\ \cmidrule(lr){3-4} \cmidrule(lr){5-6} \cmidrule(lr){7-8}
{} & {} & MAE & MSE & MAE & MSE & MAE & MSE & {} \\
\midrule
3 & {1D Poisson's equation} & 5.1e-4 \footnotemark[1] & {-} & {-} & {-} & {2.5e-7} & {8.4e-14} & {Test task: $\alpha_1=1,\alpha_2=1,\alpha_3=0.1,\alpha_4=0,\omega_1=0.7,\omega_2=1.5$, $x\in[-10,10]$} \\[0.2cm]
{} & {} & {} & {} & {} & {} & {} & {} & {NRPINN: 60 train tasks from $\alpha_1=1,\alpha_2=1,\alpha_3=0.1,\alpha_4=0,\omega_1\in[0,1],\omega_2\in[0.2]$ as per~\cite{liu2022novel}} \\[0.6cm]
{} & {} & {} & {} & {} & {} & {} & {} & {Baldwinian-PINN: 60 train tasks from $\alpha$'s$\in[0,4]$, $\omega$'s$\in[0,4]$} \\[0.5cm]
4 & {2D Poisson's equation} & 6.4e-7 \footnotemark[2] & {-} & {-} & {-} & {3.0e-7} & {2.0e-13} & {Test task: $a_1=0.15, a_2=0.18, a_3=0.20, a_4=0.31, a_5=0.43, a_6=0.56, a_7=0.70, a_8=0.80$, $b_1=0.34, b_2=0.31, b_3=0.65, b_4=0.86, b_5=0.65, b_6=0.38, b_7=0.64, b_8=0.12$, $c_1=0.84, c_2=1.07, c_3=1.12, c_4=0.83, c_5=1.12, c_6=1.11, c_7=0.99, c_8=0.91$, $d_j=0.01$, $x\in[0,1], y\in[0,1]$} \\[2.cm]
{} & {} & {} & {} & {} & {} & {} & {} & {NRPINN: 100 train tasks from $J \in \{1,5,10\}$, $a_j\in[0.1,0.9]$, $b_j\in[0.1,0.9]$, $c_j\in[0.8, 1.2]$, $d_j=0.01$ as per~\cite{liu2022novel}} \\[0.6cm]
{} & {}& {} & {} & {} & {} & {} & {} & {Baldwinian-PINN: 100 train tasks from $J\in[1,10]$, $a_j\in[-0.6,0.6]$, $b_j\in[-0.6,0.6]$, $c_j\in[0.5, 2]$, $d_j\in[0.005, 0.02]$} \\[1.cm]
5 & {Helmholtz equation} & {-} & {-} & 2.8e-2 \footnotemark[3] & {-} & {1.5e-4} & {4.7e-8} & {Test task: $\alpha_1=2.5,\alpha_2=2.5$, $x\in[-1,1], y\in[-1,1]$} \\[0.2cm]
{} & {} & {} & {} & {} & {} & {} & {} & {Hyper-LR-PINN: train tasks from $\alpha\in[2,3]$ with interval 0.1 ($\alpha=\alpha_1=\alpha_2$) as per~\cite{cho2024hypernetwork}} \\[0.6cm] 
{} & {}& {} & {} & {} & {} & {} & {} & {Baldwinian-PINN: 20 train tasks from $\alpha_1 \in [0.1,6]$, $\alpha_2 \in [0.1,6]$} \\
\end{tabular*}
\footnotetext[1]{MAE obtained after fine-tuning with 900 training iterations; result extracted from~\cite{liu2022novel}.}
\footnotetext[2]{MAE obtained after fine-tuning with 4000 training iterations; result extracted from~\cite{liu2022novel}.}
\footnotetext[3]{MAE obtained after fine-tuning with 10 training epochs; result extracted from~\cite{cho2024hypernetwork}.}
\end{sidewaystable*}

\clearpage

\section{Studies on Set of Linear ODE/PDE Problems}\label{subsec:otherlinearpde2}

The additional linear ODE/PDE problems (Problems 3-5) are described below.\newline

\subsubsection{1D Poisson's equation}

The 1D Poisson's equation consists of 60 randomly sampled train tasks $\alpha$'s $\in[0,4]$, $\omega$'s $\in[0,4]$ and 60 randomly sampled test tasks $\alpha$'s $\in[-5,5]$, $\omega$'s $\in[-5,5]$ for the PDE/BC parameters $(\alpha_1,\alpha_2,\alpha_3,\alpha_4,\omega_1,\omega_2)$:
\begin{equation}
    \text{(Problem 4)} \quad\quad\ \frac{d^2u}{dx^2} = q \quad , x\in[-10,10] \label{eq:1dpoisson}
\end{equation}
where the exact solution $u(x;\alpha_1,\alpha_2,\alpha_3,\alpha_4,\omega_1,\omega_2) =\alpha_1\mathrm{sin}(\omega_1 x) + \alpha_2\mathrm{sin}(\omega_2 x) - \alpha_3 x + \alpha_4$ is used to derive the corresponding BCs and source term $q$. 

We further test the evolved Baldwinian-PINN on a single test task ($\alpha_1=1,\alpha_2=1,\alpha_3=0.1,\alpha_4=0,\omega_1=0.7,\omega_2=1.5$) as described in~\cite{liu2022novel}.\newline

\subsubsection{2D Poisson's equation}

The 2D Poisson's equation is represented by:
\begin{subequations}
    \begin{multline} \label{eq:2dpoisson}
        \text{(Problem 5)} \quad -\left( \frac{d^2u}{dx^2} + \frac{d^2u}{dy^2} \right) = q(x,y), \\
        x \in [-1,1], \quad y \in [-1,1]
    \end{multline}
    \begin{multline} \label{eq:2dpoissonsource}
        q(x,y) = \sum_{j=1}^{J} c_j \exp \left( \frac{(x - a_j)^2 + (y - b_j)^2}{d_j} \right)
    \end{multline}
\end{subequations}
subject to BCs $u(x=-1)=0$, $u(x=1)=0$, $u(y=-1)=0$, $u(y=1)=0$. The heat source $q(x,y)$ is generated by the following scenarios: the number of heat source $J \in [1, 10]$, and coefficients sampled uniformly from $a_j\in[-0.6,0.6]$, $b_j\in[-0.6,0.6]$, $c_j\in[0.5, 2]$, and $d_j\in[0.005, 0.02]$. The training and test sets both comprise 100 tasks with different source $q(x,y)$ scenarios.

We further test the evolved Baldwinian-PINN on a single test task as described in~\cite{liu2022novel}, which is subject to a different domain $x\in[0,1], y\in[0,1]$ and BCs $u(x=0)=0$, $u(x=1)=0$, $u(y=0)=0$, $u(y=1)=0$, with $J=8$ and $a_j\in[0.1,0.9]$, $b_j\in[0.1,0.9]$, $c_j\in[0.8, 1.2]$, and $d_j=0.01$. \newline

\subsubsection{Helmholtz equation}

The Helmholtz equation consists of 20 randomly sampled train tasks and 20 randomly sampled test tasks from the PDE/BC parameters $\alpha_1 \in (0,6]$, $\alpha_2 \in (0,6]$:
\begin{multline} \label{eq:2dhelmholtz}
    \text{(Problem 6)} \quad \left( \frac{d^2u}{dx^2} + \frac{d^2u}{dy^2} \right) + u^2 = q, \\
    x \in [-1,1], \quad y \in [-1,1]
\end{multline}
where the exact solution $u(x,y;\alpha_1,\alpha_2) = \mathrm{sin}(\alpha_1 \pi x)\ \mathrm{sin}(\alpha_2 \pi y)$ is used to derive the corresponding BCs and source term $q$~\cite{cho2024hypernetwork, toloubidokhti2023dats}. 

We further test the evolved Baldwinian-PINN on a single test task ($\alpha_1=2.5,\alpha_2=2.5$) as described in~\cite{cho2024hypernetwork}.

\subsection{Results for set of linear ODE/PDE problems}

Table~\ref{tab:linearbenchmark} compares the generalization performance between Baldwinian-PINNs and baseline meta-learning PINN models (NRPINN~\cite{liu2022novel} and Hyper-LR-PINN~\cite{cho2024hypernetwork}) on linear benchmark problems.

\begin{figure*}[h]%
\centering
\includegraphics[width=0.95\textwidth]{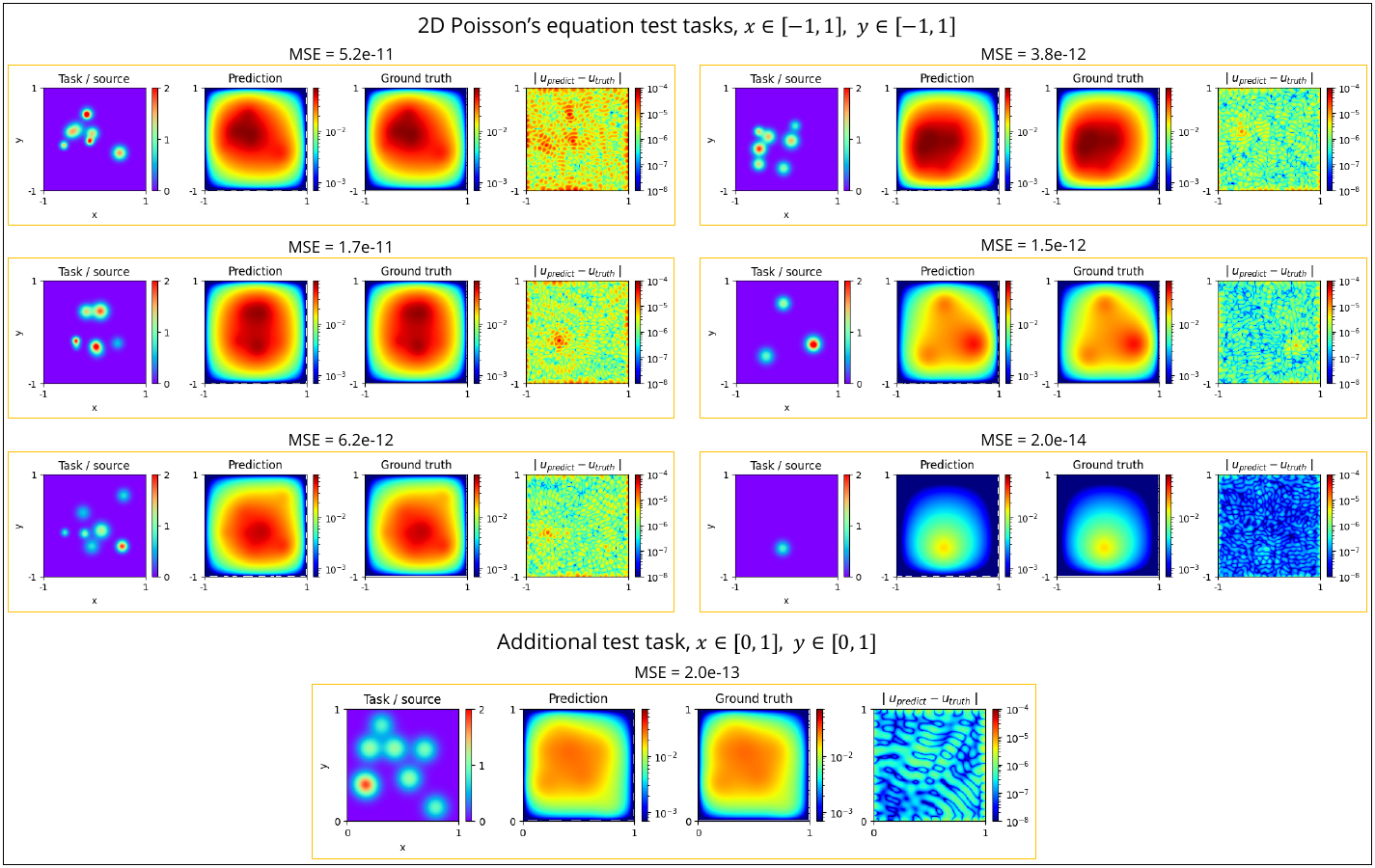}
\vspace{-0.4cm}
\caption{Baldwinian-PINN solutions selected (worst to best along the MSE spectrum) from the $n=60$ 2D Poisson’s test tasks, and an additional test task (subject to a different domain $x\in[0,1], y\in[0,1]$) described in~\cite{liu2022novel}, showing good performance on the diverse set of tasks exhibited by the source patterns.}\label{fig:additional_poisson}
\end{figure*}

\begin{figure*}[h]%
\centering
\includegraphics[width=0.95\textwidth]{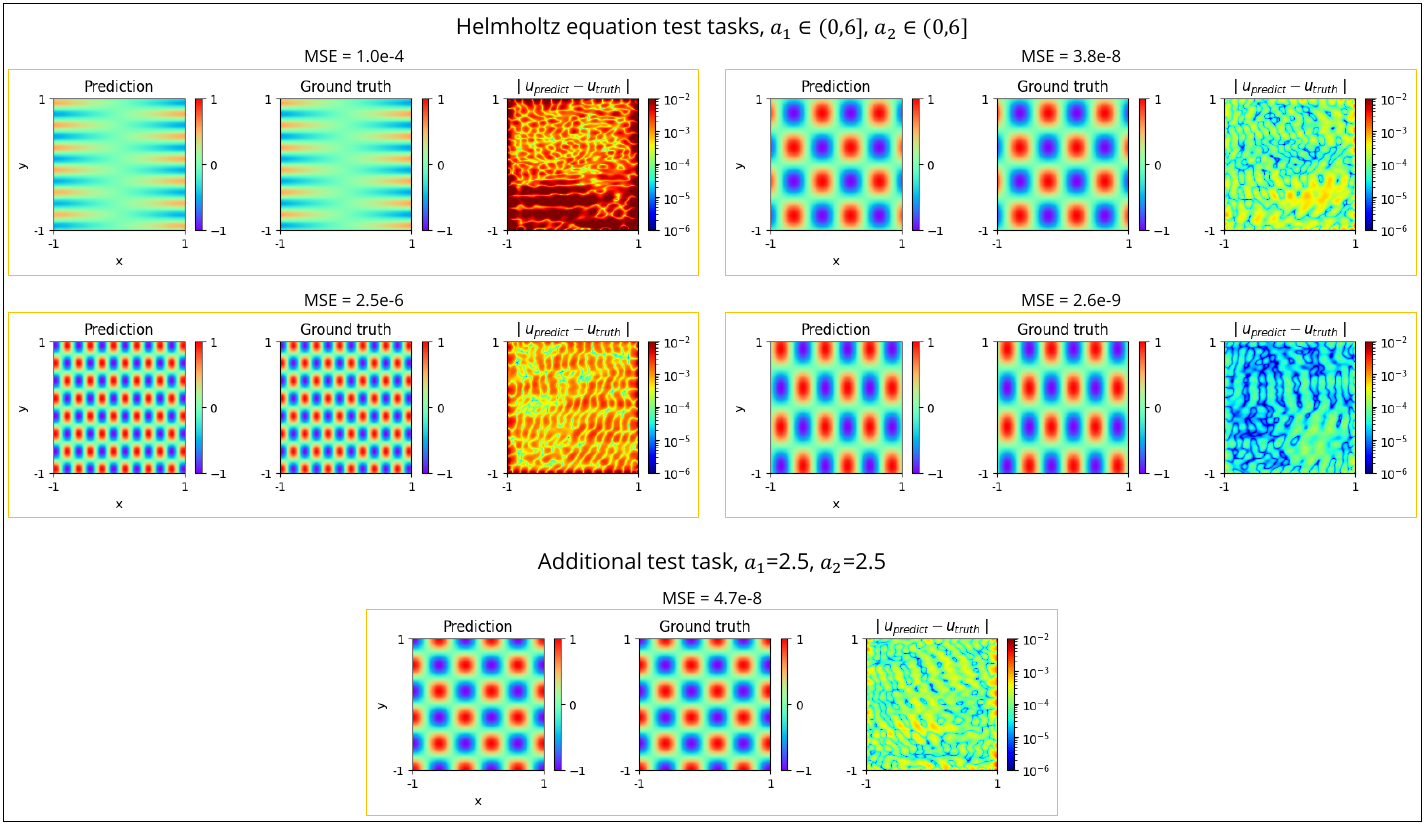}
\vspace{-0.4cm}
\caption{Baldwinian-PINN solutions selected (worst to best along the MSE spectrum) from the $n=60$ Helmholtz test tasks, and an additional test task ($\alpha_1=2.5,\alpha_2=2.5$) described in~\cite{cho2024hypernetwork}. Baldwinian-PINN shows good performance on the diverse set of PDE tasks exhibited by the frequencies.}\label{fig:additional_helmholtz}
\end{figure*}

Figure~\ref{fig:additional_poisson} and Figure~\ref{fig:additional_helmholtz} provide visualization of the Baldwinian-PINN solutions and errors for the \textit{2D Poisson’s} and \textit{Helmholtz} tasks, respectively. The visualization results show good performance on the diverse set of PDE tasks, as exhibited by source patterns and frequencies. Note that the train and test task distributions as used in this work have much greater diversity than those studied in prior meta-learning PINN works such as~\cite{liu2022novel, cho2024hypernetwork, toloubidokhti2023dats}. Our PDE parameters' range for the Helmholtz problem is 6 times larger than~\cite{cho2024hypernetwork, toloubidokhti2023dats}, thereby encapsulating a much broader frequency spectrum, including tasks in the higher frequency range which are also more challenging to learn.

To demonstrate the diversity of the tasks, we compare the performance of different Baldwinian-PINN models, with the first one learning from $n=10$ lower frequency train tasks $\alpha_1$, $\alpha_2 \in (0,1]$ for 200 neuroevolution iterations (train MSE $< 1\mathrm{e}{-11}$); while the second one is learning from $n=10$ higher frequency train tasks $\alpha_1$, $\alpha_2 \in [5,6]$ for 200 neuroevolution iterations (train MSE $< 1\mathrm{e}{-6}$). Both low- and high-frequency-learned models are applied to the same set of $n=60$ test tasks drawing from the full frequency range $\alpha_1$, $\alpha_2 \in (0,6]$, and their test MSE results are 4.8$\mathrm{e}{-2}$ {\tiny$\pm$1.5$\mathrm{e}{-1}$} and 3.8$\mathrm{e}{-2}$ {\tiny$\pm$8.7$\mathrm{e}{-2}$}, respectively.

Their test MSE results are 2-3 orders of magnitudes higher than the 1.6$\mathrm{e}{-5}$ {\tiny$\pm$5.1$\mathrm{e}{-5}$} achieved by the original Baldwinian-PINN. Note that the original model is learned from $n=20$ train tasks $\alpha_1$, $\alpha_2 \in (0,6]$ for 400 neuroevolution iterations (train MSE $< 1\mathrm{e}{-5}$).

\begin{figure*}[h]%
\centering
\includegraphics[width=1.0\textwidth]{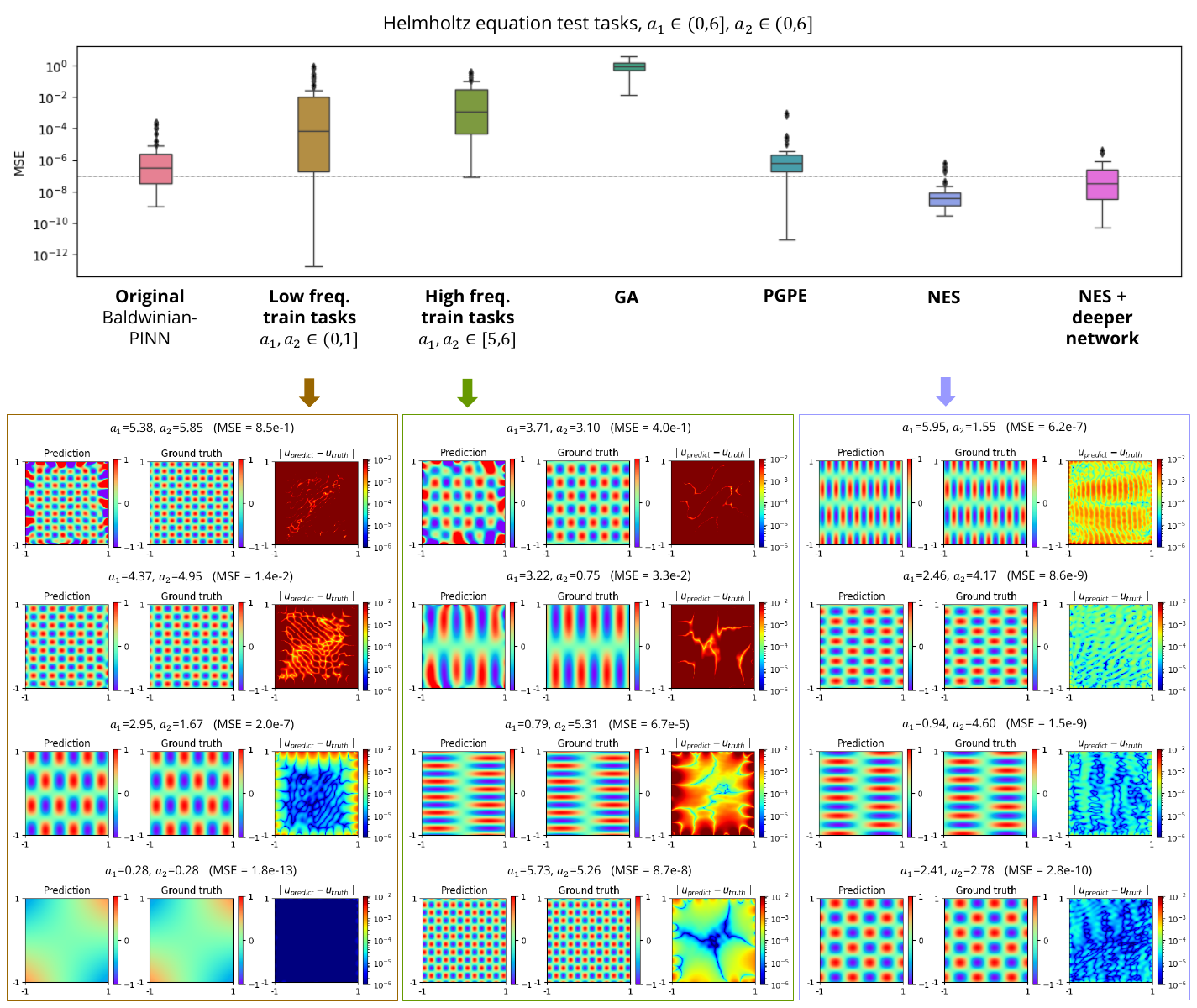}
\vspace{-0.4cm}
\caption{The MSE distributions on $n=60$ Helmholtz test tasks for the original Baldwinian-PINN (evolved by NES to meta-learn from $n=20$ train tasks $\alpha_1 \in (0,6]$, $\alpha_2 \in (0,6]$) and its variants, i.e., using a different set of train tasks, or a different evolutionary algorithm, or both different evolutionary algorithm and neural architecture. The solutions and errors for selected (worst to best along the MSE spectrum) Helmholtz test tasks are visualized for both low- and high-frequency-learned Baldwinian-PINN models, and also the NES-evolved Baldwinian-PINN.}\label{fig:additional_helmholtz_models}
\end{figure*}

Figure~\ref{fig:additional_helmholtz_models} provides the comparison of test MSE distributions and visualization of selected (worst to best along the MSE spectrum) Baldwinian-PINN solutions and errors for the low- and high-frequency-learned Balwinian-PINN models. From the results, we can observe that the low-frequency-learned model tends to give inaccurate physics-informed predictions for the test tasks from the other side of the frequency spectrum (i.e., high and mixed frequencies), and vice versa.

These results highlight the effectiveness of the Baldwinian neuroevolution for generating Baldwinian-PINN models that are ``genetically equipped" to perform well over diverse task distribution pertaining to the training environment, e.g., low frequency, high frequency, or full frequency spectrum, as represented in the population.


\begin{figure*}[h]%
\centering
\includegraphics[width=1.0\textwidth]{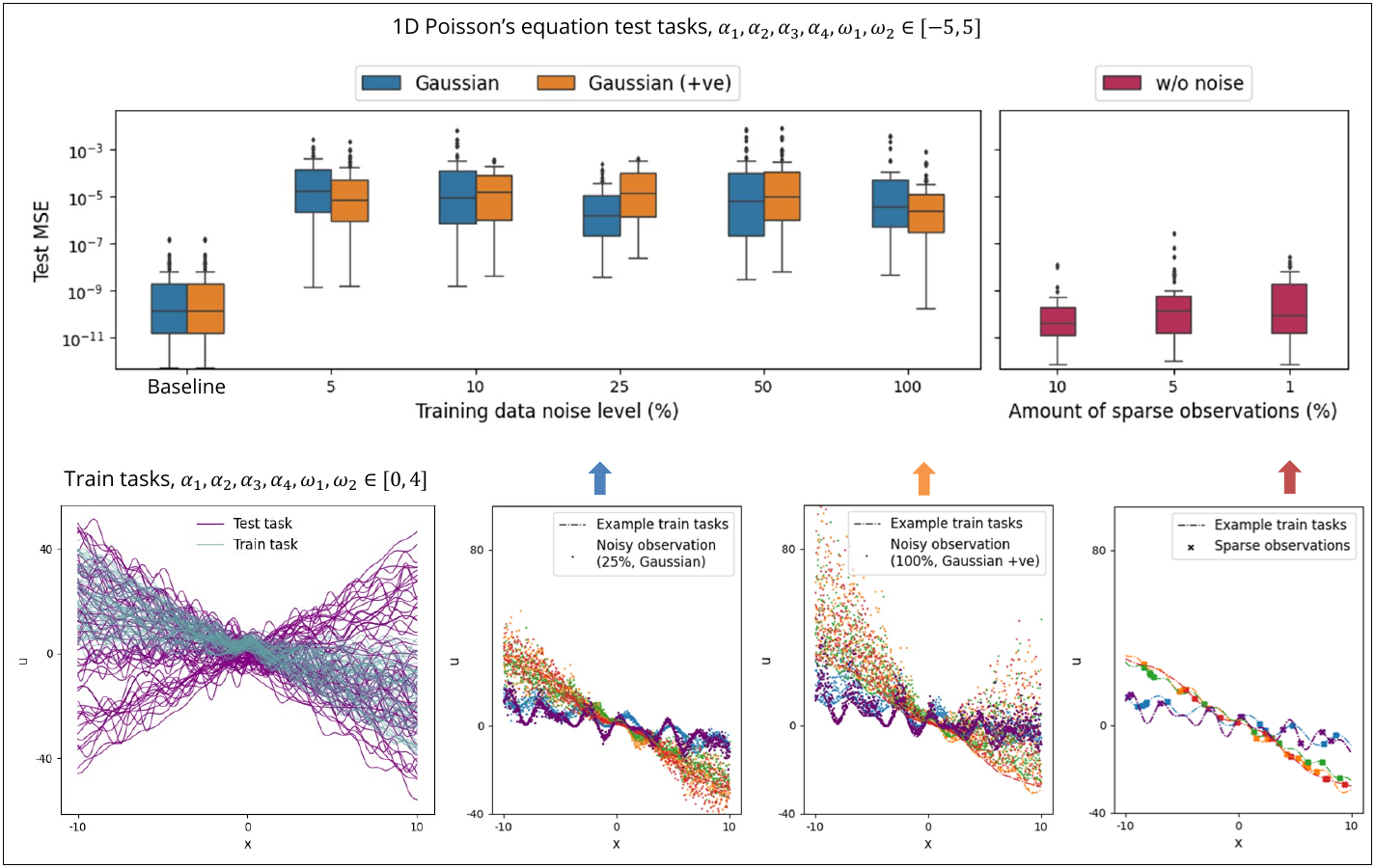}
\vspace{-0.4cm}
\caption{Robustness to noisy and sparse observations on 1D Poisson’s problem. Top: distribution of test MSE obtained by Baldwinian-PINNs when noisy or sparse training-task labels are used. The training data noise level ranges from 0\% (baseline) to 100\% for both zero-mean “Gaussian” noise and positively biased “Gaussian (+ve)” noise scenarios, whereas the number of sparse observations ranges from 100\% (baseline) to 1\%. Bottom: examples of training tasks with corresponding noisy (25\% Gaussian; 100\% Gaussian +ve) and sparse (1\%) observation scenarios. Baldwinian-PINNs degrade gracefully with noise and exhibit low error with very limited observations.}\label{fig:additional_noise}
\end{figure*}


\subsection{Baldwinian-PINN with different evolutionary algorithms and neural architectures} \label{subsec:additional_evo_depth}

The CMA-ES algorithm used in our experimental study is merely an instantiation of the proposed Baldwinian neuroevolution framework for meta-learning PINN. The outer-loop evolution procedure is agnostic to, and can be seamlessly switched to other state-of-the-art evolutionary search algorithms.

In this section, we present the results when Baldwinian neuroevolution is carried out by 3 alternate algorithms: simple genetic algorithm (GA)~\cite{evojax2022}, policy gradients with parameter-based exploration (PGPE)~\cite{sehnke2010parameter}, and natural evolution strategies (NES) variant~\cite{nomura2022fast}, using the \textit{Helmholtz} example. Note that the same optimization settings (e.g., initial standard deviation, population size, neuroevolution iterations) as the original (CAM-ES-evolved) Baldwinian-PINN are used (i.e., we do not fine-tune the settings for each algorithm). The search dimension consists of 36 network hyperparameters controlling the distributional mean and spread of the weights and biases, the learning hyperparameter $\lambda$, and the loss importance hyperparameter $\lambda_{pde}$ for re-balancing the relative importance between PDE and BC/IC errors in the least-squares solution.

Their test MSE values are 1.1 {\tiny$\pm$8.1$\mathrm{e}{-1}$}, 3.0$\mathrm{e}{-5}$ {\tiny$\pm$1.5$\mathrm{e}{-4}$}, and 2.9$\mathrm{e}{-8}$ {\tiny$\pm$9.8$\mathrm{e}{-8}$}, respectively. The simple GA performs much worse than the original Baldwinian-PINN. On the other hand, the NES-evolved Baldwinian-PINN significantly outperforms the CMA-ES-evolved Baldwinian-PINN, with more than 2 orders of magnitude MSE improvement on test tasks. Their MSE distributions on $n=60$ test tasks are compared in Figure~\ref{fig:additional_helmholtz_models}, together with visualization of the selected (worst to best along the MSE spectrum) solutions and errors obtained by the NES-evolved Baldwinian-PINN.

Similarly, the Baldwinian-PINN neural architecture described in Section~\ref{subsec:architecture} is merely an instantiation of the proposed Baldwinian neuroevolution framework for meta-learning PINN. The extension from a simple, single hidden layer neural architecture to a deeper, more complex neural architecture is straightforward. 

To demonstrate this, we construct a MLP with multiple hidden layers with a mix of \textit{sin} and \textit{softplus} activation functions, and additional skip connections from the early hidden layers to the final layer. The total number of weight parameters in this deeper Baldwinian-PINN is 68,480 (including 1280 weights in the final layer). Note that the learned final layer is the outcome of lifetime learning through the Tikhonov regularization computation, whereas these 67,200 weight parameters before the final layer are sampled from 32 normal distributions. There are 64 distributional hyperparameters, 1 learning hyperparameter $\lambda$, and 1 loss importance hyperparameter $\lambda_{pde}$ evolved by the NES algorithm, using the same optimization settings (e.g., initial standard deviation, population size, neuroevolution iterations) as the original Baldwinian-PINN. For this study, the original Baldwinian-PINN architecture consists of 3600 weight parameters (including 900 weights in the final layer).

Despite having a much larger number of weight parameters in the deeper Baldwinian-PINN model, Baldwinian neuroevolution managed to arrive at a good set of distributional hyperparameters for the model to make a fast and accurate physics-based prediction on the test tasks (test MSE=2.6$\mathrm{e}{-7}$ {\tiny$\pm$6.7$\mathrm{e}{-7}$}), in the Helmholtz problem.

The results described in this section underscore the promise of the Baldwinian neuroevolution framework for meta-learning PINN. Further improvements to the search algorithm and neural architecture design may boost the Baldwinian-PINN's learning speed and accuracy on other complex physics problems. 

\subsection{Noisy or scarce training-task labels}\label{subsec:additional_noise}

Our lifetime-learning step is entirely physics-based, requires no labels at test time, and is solved in closed form with Tikhonov regularization; this inherently resists noisy or scarce data. When labels in the training tasks are noisy or sparse, we can rebalance the physics-data trade-off in the loss function by tuning the loss weights $\kappa_{LSE}$ and $\kappa_{MSE}$ to further strengthen the physics prior.

Empirically, Figure~\ref{fig:additional_noise} evaluates Baldwinian-PINNs’ lifetime-learning performance on a set of 1D Poisson’s equation tasks ($n=60$) when the training-task labels are noisy or scarce. As observation noise increases from 0 to 100\% under both zero-mean Gaussian noise and positively biased noise scenarios, the test error rises smoothly rather than catastrophically. Meanwhile, test accuracy remains low even when only 1–10\% of points from the training tasks are observed, because the physics residual and BC/IC rows constrain the solution and meta-learning can be effectively guided by a few quality measurements, demonstrating that Baldwinian-PINNs tolerate non-ideal data and can operate under severe data scarcity.

\clearpage

\begin{sidewaystable*}
\caption{Comparison with baseline meta-learning PINN models on nonlinear benchmark problems.} \label{tab:benchmark}
\begin{tabular*}{\textheight}{@{\extracolsep\fill}p{.1cm}>{\raggedright\arraybackslash}p{3cm}p{1cm}p{.5cm}p{1cm}p{.5cm}p{1cm}p{.5cm}p{1cm}p{.5cm}p{1cm}p{.5cm}p{1cm}p{.5cm}p{1cm}p{.5cm}p{1cm}p{.5cm}}
\toprule
{} & \textbf{Problem} & \multicolumn{16}{@{}c@{}}{\textbf{Method / Model}}\\ \cmidrule(lr){3-18}
{} & {(no. test task)} & \multicolumn{2}{@{}c@{}}{Random} & \multicolumn{2}{@{}c@{}}{MAML} & \multicolumn{2}{@{}c@{}}{Center} & \multicolumn{2}{@{}c@{}}{Multitask} & \multicolumn{2}{@{}c@{}}{LMC} & \multicolumn{2}{@{}c@{}}{RBF (multiquadric)} & \multicolumn{2}{@{}c@{}}{Polynomial} & \multicolumn{2}{@{}c@{}}{Baldwinian-PINN} \\ \cmidrule(lr){3-4} \cmidrule(lr){5-6} \cmidrule(lr){7-8} \cmidrule(lr){9-10} \cmidrule(lr){11-12} \cmidrule(lr){13-14} \cmidrule(lr){15-16} \cmidrule(lr){17-18}
{} & {} & Error\footnotemark[1] & Time\footnotemark[2] & Error\footnotemark[1] & Time\footnotemark[2] & Error\footnotemark[1] & Time\footnotemark[2] & Error\footnotemark[1] & Time\footnotemark[2] & Error\footnotemark[1] & Time\footnotemark[2] & Error\footnotemark[1] & Time\footnotemark[2] & Error\footnotemark[1] & Time\footnotemark[2] & Error\footnotemark[3] & Time\footnotemark[4] \\
\midrule
7 & Burgers’ equation $(n=32)$ & 1.2e-3 {\tiny$\pm$1.8e-3} & 105 {\tiny$\pm$29} & 1.7e-3 {\tiny$\pm$3.2e-3} & 125 {\tiny$\pm$38} & 9e-4 {\tiny$\pm$5e-4} & 56 {\tiny$\pm$48} & 7e-4 {\tiny$\pm$2e-4} & 24 {\tiny$\pm$21} & 8e-4 {\tiny$\pm$4e-4} & 40 {\tiny$\pm$27} & 8e-4 {\tiny$\pm$4e-4} & 7 {\tiny$\pm$8} & 7e-4 {\tiny$\pm$3e-4} & 7 {\tiny$\pm$16} & \textbf{3.8e-4} {\tiny$\pm$6.9e-4} & \textbf{1.96} {\tiny$\pm$0.01} \\[0.1cm]
8 & nonlinear heat $(n=64)$ & 5.2e-3 {\tiny$\pm$3.9e-3} & 156 {\tiny$\pm$42} & 4.5e-3 {\tiny$\pm$3.1e-3} & 188 {\tiny$\pm$43} & 4.5e-3 {\tiny$\pm$2.4e-3} & 96 {\tiny$\pm$41} & 4.8e-3 {\tiny$\pm$3.6e-3} & 49 {\tiny$\pm$30} & 4.9e-3 {\tiny$\pm$3.3e-3} & 59 {\tiny$\pm$28} & 4.4e-3 {\tiny$\pm$2.2e-3} & 35 {\tiny$\pm$30} & 4.6e-3 {\tiny$\pm$3.2e-3} & 38 {\tiny$\pm$30} & \textbf{6.0e-4} {\tiny$\pm$7.6e-4} & \textbf{0.87} {\tiny$\pm$0.02} \\[0.1cm]
9 & Allen-Cahn equation $(n=32)$ & 1.5e-2 {\tiny$\pm$1.2e-2} & 496 {\tiny$\pm$161} & {-} & {-} & 1.2e-2 {\tiny$\pm$4.6e-3} & 201 {\tiny$\pm$83} & 1.2e-2 {\tiny$\pm$4.7e-3} & 120 {\tiny$\pm$44} & 1.1e-2 {\tiny$\pm$4.0e-3} & 120 {\tiny$\pm$21} & 1.2e-2 {\tiny$\pm$5.3e-3} & 68 {\tiny$\pm$46} & 1.2e-2 {\tiny$\pm$4.2e-3} & 44 {\tiny$\pm$19} & \textbf{9.5e-4} {\tiny$\pm$1.5e-3} & \textbf{1.19} {\tiny$\pm$0.01} \\[0.1cm]
10 & Diffusion-reaction $(n=64)$ & 1.1e-2 {\tiny$\pm$6.2e-3} & 1073 {\tiny$\pm$206} & {-} & {-} & 9.5e-3 {\tiny$\pm$5.9e-3} & 426 {\tiny$\pm$159} & 9.0e-3 {\tiny$\pm$5.4e-3} & 243 {\tiny$\pm$93} & 9.1e-3 {\tiny$\pm$5.8e-3} & 302 {\tiny$\pm$127} & 8.1e-3 {\tiny$\pm$4.6e-3} & 280 {\tiny$\pm$161} & 8.5e-3 {\tiny$\pm$5.1e-3} & 249 {\tiny$\pm$129} & \textbf{1.3e-4} {\tiny$\pm$2.0e-4} & \textbf{0.58} {\tiny$\pm$0.02} \\[0.1cm]
11 & 6D diffusion-reaction $(n=100)$ & 2.2e-3 {\tiny$\pm$1.3e-3} & 612 {\tiny$\pm$255} & {-} & {-} & 1.8e-3 {\tiny$\pm$1.7e-3} & 494 {\tiny$\pm$222} & 1.7e-3 {\tiny$\pm$1.9e-3} & 431 {\tiny$\pm$200} & 1.5e-3 {\tiny$\pm$1.8e-3} & 428 {\tiny$\pm$201} & 1.7e-3 {\tiny$\pm$2.0e-3} & 375 {\tiny$\pm$168} & 1.8e-3 {\tiny$\pm$2.0e-3} & 496 {\tiny$\pm$213} & \textbf{1.9e-4} {\tiny$\pm$1.5e-4} & \textbf{0.60} {\tiny$\pm$0.03} \\
\end{tabular*}
\footnotetext{Benchmark results from meta-learning PINN method / model (i.e., Random, MAML, Center, Multitask, LMC, RBF (multiquadric), Polynomial) are extracted from \cite{penwarden2023metalearning}.}
\footnotetext[1]{Relative L2 errors are obtained from a meta-learned PINN after fine-tuning with ADAM (500 iterations) + L-BFGS optimizations.}
\footnotetext[2]{Only for the L-BFGS optimization time (s) of the fine-tuning phase (computation time of meta-learning phase, and ADAM optimization prior to the L-BFGS during fine-tuning phase not reported in \cite{penwarden2023metalearning})}.
\footnotetext[3]{Relative L2 errors are aggregated from 5 individual runs.}
\footnotetext[4]{Computation time (s) per task during lifetime-learning (fine-tuning) phase, on single GPU (NVIDIA GeForce RTX 3090). For neuroevolution time please refer to Table~\ref{tab:summary}}
\end{sidewaystable*}

\clearpage

\section{Studies on Set of Nonlinear PDE Problems}\label{subsec:nonlinearpde2}

The nonlinear PDE problems (Problems 7-11) used for demonstrating the Baldwinian neuroevolution of physics are described below.\newline

\subsubsection{Burgers’ equation}

The Burgers’ equation~\cite{bec2007burgers} consists of 16 randomly sampled train tasks and 32 uniformly sampled test tasks from the PDE parameter $\gamma\in[5\mathrm{e}{-3},5\mathrm{e}{-2}]$:
\begin{multline} \label{eq:burgers}
    \text{(Problem 7)} \quad \frac{du}{dt} + u \frac{du}{dx} - \gamma \frac{d^2u}{dx^2} = 0, \\
    x \in [-1,1], \quad t \in (0,1]
\end{multline}
subject to IC $u(x,t=0) = -\mathrm{sin}(\pi x)$. \newline

\subsubsection{Nonlinear heat equation}

The nonlinear heat equation consists of 13 randomly sampled train tasks and 64 uniformly sampled test tasks from the PDE parameters $\gamma\in[1,\pi],k\in[1,\pi]$:
\begin{multline} \label{eq:heat}
    \text{(Problem 8)} \quad\quad \frac{du}{dt} - \gamma \frac{d^2u}{dx^2} + k\ \mathrm{tanh}(u) = q, \\
    x\in[-1,1], \quad t\in(0,1]    
\end{multline}
where the exact solution $u(x,t;\gamma,k) = k\mathrm{sin}(\pi x) \mathrm{exp}(-\pi kx^2) \mathrm{exp}(-\pi t^2)$ is used to derive the corresponding IC, BCs and source term $q$. \newline

\subsubsection{Nonlinear Allen-Cahn equation}

The nonlinear Allen-Cahn equation consists of 16 randomly sampled train tasks and 32 uniformly sampled test tasks from the PDE parameter $\gamma\in(0,\pi]$:
\begin{multline} \label{eq:allencahn}
    \text{(Problem 9)} \quad\quad \gamma (\frac{d^2u}{dx^2} + \frac{d^2u}{dy^2}) + u(u^2-1) = q, \\
    x\in[-1,1], \quad y\in[-1,1] 
\end{multline}

where the exact solution $u(x,y;\gamma) = \mathrm{exp}(-\gamma (x+0.7)) \mathrm{sin}(\pi x) \mathrm{sin}(\pi y)$ is used to derive the corresponding BCs and source term $q$. \newline

\subsubsection{Nonlinear diffusion-reaction equation}

The nonlinear diffusion-reaction equation~\cite{rao2023encoding} consists of 22 randomly sampled train tasks and 64 uniformly sampled test tasks from the PDE parameter $\gamma\in[1,\pi],k\in[1,\pi]$:
\begin{multline} \label{eq:diffusionreaction}
    \text{(Problem 10)} \quad \gamma \left( \frac{d^2u}{dx^2} + \frac{d^2u}{dy^2} \right) + k u^2 = q, \\
    x \in [-1,1], \quad y \in [-1,1]
\end{multline}
where the exact solution $u(x,y;\gamma,k) = k\mathrm{sin}(\pi x) \mathrm{sin}(\pi y) \mathrm{exp}(-\gamma\sqrt{kx^2+y^2})$ is used to derive the corresponding BCs and source term $q$. \newline

\subsubsection{6D parametric diffusion-reaction}

The 6D parametric diffusion-reaction problem consists of 17 randomly sampled train tasks and 100 randomly sampled test tasks from the PDE/BC parameters $(\alpha_1,\alpha_2,\omega_1,\omega_2,\omega_3,\omega_4)$, $\alpha$'s$\in[0.1,1]$, $\omega$'s$\in[1,5]$:
\begin{multline} \label{eq:6ddiffusionreaction}
    \text{(Problem 11)} \quad \left( \frac{d^2u}{dx^2} + \frac{d^2u}{dy^2} \right) + u(1 - u^2) = q, \\
    x \in [-1,1], \quad y \in [-1,1]
\end{multline}
where the exact solution $u(x,y;\alpha_1,\alpha_2,\omega_1,\omega_2,\omega_3,\omega_4) =\alpha_1\mathrm{tanh}(\omega_1 x)\mathrm{tanh}(\omega_2 y) + \alpha_2\mathrm{sin}(\omega_3 x)\mathrm{sin}(\omega_4 y) $ is used to derive the corresponding BCs and source term $q$.

\subsection{Results for set of nonlinear PDE problems} \label{subsec:additional_nonlinearpde}

As per \cite{penwarden2023metalearning}, the spatio-temporal domain in Problems 7-8 is uniformly discretized into 256$\times$100, and the 2D spatial domain in Problems 9-11 is uniformly discretized into 128$\times$128, for the test tasks. The comparison of generalization performance and compute cost (i.e., lifetime learning for Baldwinian-PINNs or gradient fine-tuning for baselines) between Baldwinian-PINN and several meta-learned PINN models (based on results reported in \cite{penwarden2023metalearning}) are summarized in Table~\ref{tab:benchmark}.

\begin{figure*}[h]%
\centering
\includegraphics[width=1.0\textwidth]{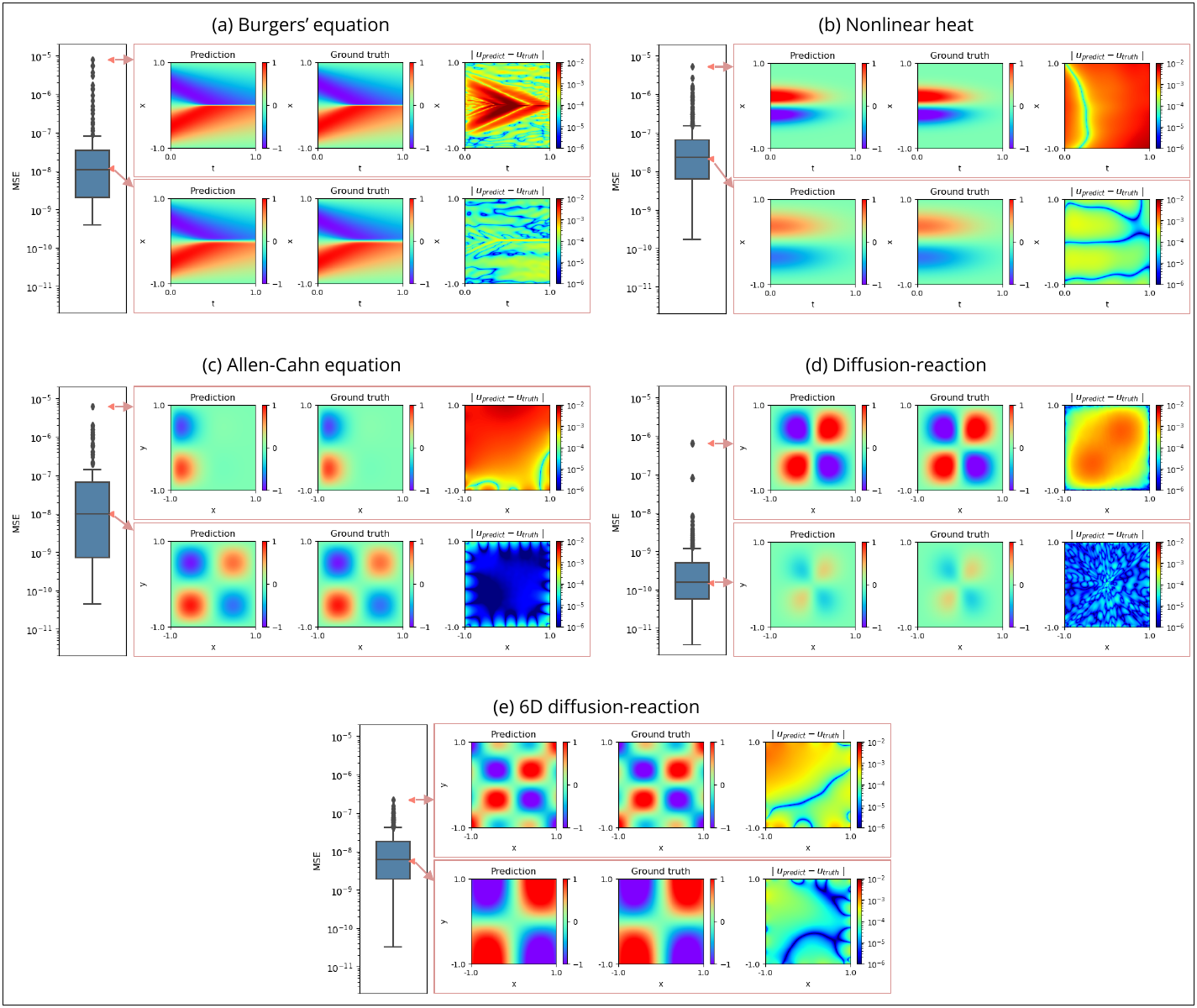}
\vspace{-0.4cm}
\caption{Baldwinian-PINN solutions with the worst and (approximately) average accuracy along the MSE spectrum (pooled from all test tasks $\times$ 5 individual runs), for 5 nonlinear PDE problems.}\label{fig:additional_nonlinear}
\end{figure*}

Figure~\ref{fig:additional_nonlinear} provides additional visualization results for the 5 nonlinear PDE problems described above, showing the corresponding Baldwinian-PINN solutions with the worst and median accuracy along the MSE spectrum (pooled from all test tasks $\times$ 5 individual runs) for each problem.


\section{Additional Results for Navier-Stokes (N-S) Equations for Fluid Flow} \label{subsec:additional_nsflow}

\begin{figure*}[h]%
\centering
\includegraphics[width=1.0\textwidth]{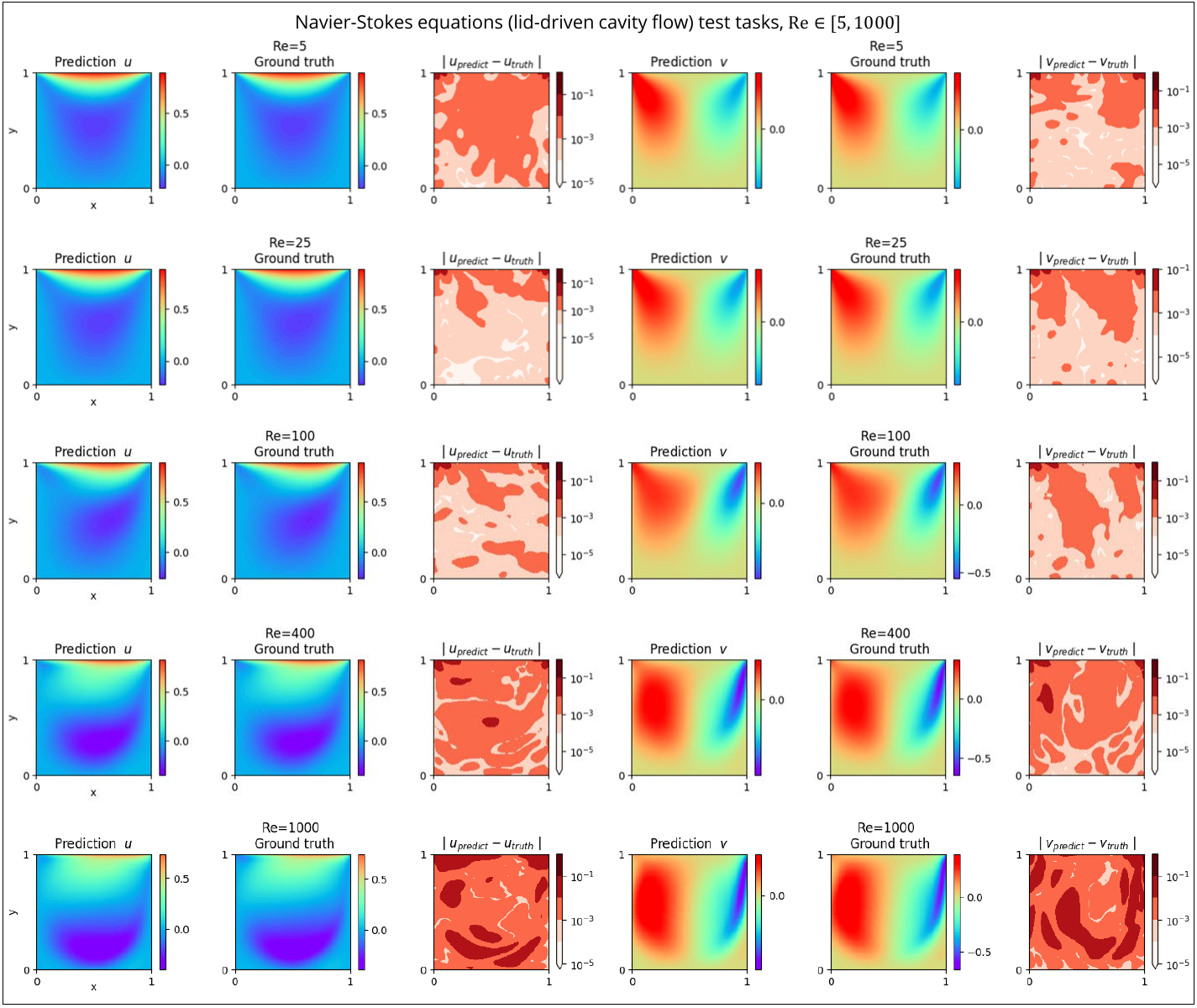}
\vspace{-0.4cm}
\caption{Baldwinian-PINN prediction vs. ground truth and their absolute error maps for unseen fluid flow (N-S equations) tasks on $Re\in[5,1000]$. The MSE of the solutions obtained by the best evolved Baldwinian-PINN (sampled from the center of NES search distribution after 400 iterations with initial std. = 0.5) compared to numerical simulation ground truths are 4.4$\mathrm{e}{-5}$ (Re=5), 3.4$\mathrm{e}{-5}$ (Re=25), 3.5$\mathrm{e}{-5}$ (Re=100), 5.1$\mathrm{e}{-5}$ (Re=400), and 1.8$\mathrm{e}{-4}$ (Re=1000), respectively.}\label{fig:additional_nsflow}
\end{figure*}

Our Baldwinian-PINN model for the fluid flow problem consists of multiple hidden layers with a mix of \textit{sin}, \textit{tanh}, and \textit{solfplus} activation functions, with skip connections from the early hidden layers to the final layer. The total number of weight parameters is 133,504 (including 1152 weights in the final layer). Figure~\ref{fig:additional_nsflow} provides additional visualization comparing Baldwinian-PINN predictions and ground truth for the lid-driven cavity flow across unseen Reynolds numbers $Re=\{5,25,100,400,1000\}$, showing $u$- and $v$-velocity fields alongside absolute error maps. Errors remain uniformly small across the domain and over two orders of magnitude in $Re$, indicating strong generalization of the Baldwinian-PINN model to markedly different flow regimes.


\clearpage


\begin{sidewaystable*}
\caption{DNN / PINN model and training configurations used in ablation study.} \label{tab:baseline_model}
\begin{tabular*}{\textheight}{@{\extracolsep\fill}lclllllp{1.35cm}ll}
\toprule
{} & \textbf{Problem} & \textbf{Model} & \textbf{Architecture} & \textbf{Activation} & \textbf{Initialization} & \textbf{No. weights} & \multicolumn{3}{@{}c@{}}{\textbf{Training / ADAM optimizer}} \\ \cmidrule(lr){8-10}%
{} & {} & {} & {} & {} & {} & {(final layer)} & {Batch size for task} & {Max. iteration} & {Learning rate\footnotemark[1]} \\
\midrule
\multirow{4}{*}{\centering 1} & \multirow{4}{*}{\centering \makecell{Convection-diffusion \\
$(x,\alpha)\rightarrow u$}} & {1a} & {Deep} & {tanh} & {Xavier} & {2750 (50)} & {10} & {50,000} & {5e-4 / 5e-3 / 5e-2 } \\
{} & {} & {1b} & {Deep} & {sin} & {He} & {2750 (50)} & {10} & {50,000} & {5e-4 / 5e-3 / 5e-2 } \\
{} & {} & {2a} & {Shallow} & {tanh} & {Xavier} & {3600 (900)} & {10} & {50,000} & {5e-4 / 5e-3 / 5e-2 } \\
{} & {} & {2b} & {Shallow} & {sin} & {He} & {3600 (900)} & {10} & {50,000} & {5e-4 / 5e-3 / 5e-2 } \\
\midrule
\multirow{4}{*}{\centering 6} & \multirow{4}{*}{\centering \makecell{Kinematics \\
$(t, vel_0, a_0, C_d, A, m)\rightarrow (x,y)$}} & {1a} & {Deep} & {tanh} & {Xavier} & {3930 (60)} & {30} & {150,000} & {5e-4 / 5e-3 / 5e-2} \\
{} & {} & {1b} & {Deep} & {sin} & {He} & {3930 (60)} & {30} & {150,000} & {5e-4 / 5e-3 / 5e-2} \\
{} & {} & {2a} & {Shallow} & {tanh} & {Xavier} & {6300 (1800)} & {30} & {150,000} & {5e-4 / 5e-3 / 5e-2} \\
{} & {} & {2b} & {Shallow} & {sin} & {He} & {6300 (1800)} & {30} & {150,000} & {5e-4 / 5e-3 / 5e-2} \\
\end{tabular*}
\footnotetext[1]{Flat learning rate for first 40\% of iterations, then cosine decay towards 1e-6.}
\end{sidewaystable*}

\clearpage


\begin{sidewaystable*}
\caption{Generalization performance of DNN / PINN models in ablation study.} \label{tab:baseline_results}
\begin{tabular*}{\textheight}{@{\extracolsep\fill}lccp{1cm}p{1cm}p{1cm}p{1cm}p{1cm}p{1cm}p{1cm}p{1cm}p{1cm}p{1cm}p{1cm}p{1cm}p{1cm}}
\toprule
{} & \textbf{Problem} & \textbf{Model} & {} & \multicolumn{6}{@{}c@{}}{\textbf{DNN}} & \multicolumn{6}{@{}c@{}}{\textbf{PINN}}\\
\cmidrule(lr){5-10} \cmidrule(lr){11-16}
{} & {(no. test task)} & {} & {Lr\footnotemark[1]} & \multicolumn{2}{@{}c@{}}{5e-4} & \multicolumn{2}{@{}c@{}}{5e-3} & \multicolumn{2}{@{}c@{}}{5e-2} & \multicolumn{2}{@{}c@{}}{5e-4} & \multicolumn{2}{@{}c@{}}{5e-3} & \multicolumn{2}{@{}c@{}}{5e-2} \\
\cmidrule(lr){5-6} \cmidrule(lr){7-8} \cmidrule(lr){9-10} \cmidrule(lr){11-12} \cmidrule(lr){13-14} \cmidrule(lr){15-16}
{} & {} & {} & {Mode\footnotemark[2]} & DNN & Tik-Reg & DNN & Tik-Reg & DNN & Tik-Reg & PINN & Tik-Reg & PINN & Tik-Reg & PINN & Tik-Reg \\
\midrule
\multirow{8}{*}{\centering 1} & \multirow{8}{*}{\centering \makecell{Convection-diffusion \\ $(x,\alpha)\rightarrow u$ \\ \\ $(n=110)$}} & 1a & {} & 3.5e-4 {\tiny$\pm$ 2.2e-3} & 1.5e-1 {\tiny$\pm$ 1.1e-1} & \cellcolor{lime!50} 2.3e-4 {\tiny$\pm$1.8e-3} & \cellcolor{lime!50} 3.8e-2 {\tiny$\pm$7.6e-2} & 1.2e-2 {\tiny$\pm$2.6e-2} & 2.3e-1 {\tiny$\pm$2.8e-2} & 4.6e-3 {\tiny$\pm$2.0e-2} & 1.4e-2 {\tiny$\pm$4.3e-2}  & \cellcolor{cyan!50} 1.9e-3 {\tiny$\pm$1.1e-2}  & \cellcolor{cyan!50} 7.2e-3 {\tiny$\pm$2.5e-2} & 2.2e-1 {\tiny$\pm$5.6e-2} & 2.3e-1 {\tiny$\pm$2.8e-2} \\[0.15cm]
{} & {} & 1b & {} & 6.9e-4 {\tiny$\pm$2.2e-3} & 1.8e-1 {\tiny$\pm$1.0e-1} & 1.7e-4 {\tiny$\pm$1.0e-3} & 1.3e-1 {\tiny$\pm$1.2e-1} & 1.9e-3 {\tiny$\pm$7.4e-3} & 2.2e-2 {\tiny$\pm$3.6e-2} & 1.3e-1 {\tiny$\pm$9.3e-2} & 1.6e-1 {\tiny$\pm$1.1e-1} & 2.1e-2 {\tiny$\pm$3.5e-2} & 1.1e-1 {\tiny$\pm$1.1e-1} & 3.4e-1 {\tiny$\pm$3.7e-1} & 1.9e-1 {\tiny$\pm$9.6e-2} \\[0.15cm]
{} & {} & 2a & {} & 7.1e-3 {\tiny$\pm$6.8e-3} & 2.3e-1 {\tiny$\pm$4.7e-2} & 4.5e-3 {\tiny$\pm$1.3e-2} & 2.2e-1 {\tiny$\pm$5.3e-2} & 2.3e-3 {\tiny$\pm$4.0e-3} & 2.1e-1 {\tiny$\pm$7.2e-2} & 2.2e-1 {\tiny$\pm$4.5e-2} & 2.3e-1 {\tiny$\pm$4.9e-2} & 1.9e-1 {\tiny$\pm$7.0e-2} & 2.2e-1 {\tiny$\pm$5.6e-2} & 2.0e-1 {\tiny$\pm$6.5e-2}  & 2.1e-1 {\tiny$\pm$7.8e-2} \\[0.15cm]
{} & {} & 2b & {} & 6.0e-3 {\tiny$\pm$7.1e-3} & 2.2e-1 {\tiny$\pm$5.7e-2} & 3.8e-3 {\tiny$\pm$5.0e-3} & 2.2e-1 {\tiny$\pm$6.1e-2} & 2.5e-3 {\tiny$\pm$5.2e-3} & 2.1e-1 {\tiny$\pm$7.7e-2} & 2.2e-1 {\tiny$\pm$5.1e-2} & 2.2e-1 {\tiny$\pm$5.8e-2} & 2.2e-1 {\tiny$\pm$5.2e-2} & 2.2e-1 {\tiny$\pm$5.9e-2} & 2.1e-1 {\tiny$\pm$5.8e-2} & 2.2e-1 {\tiny$\pm$6.1e-2} \\[0.1cm]
\midrule
\multirow{8}{*}{\centering 6} & \multirow{8}{*}{\centering \makecell{Kinematics \\ $(t, vel_0, a_0, C_d, A, m)\rightarrow (x,y)$ \\ \\ $(n=100)$}} & 1a & {} & 8.5e+0 {\tiny$\pm$2.8e+1} & 2.9e-2 {\tiny$\pm$4.2e-1} & 3.7e+1 {\tiny$\pm$8.8e+1} & 9.4e+2 {\tiny$\pm$3.5e+3} & 2.0e+2 {\tiny$\pm$6.2e+2} & 2.6e+3 {\tiny$\pm$4.6e+3} & 1.1e+1 {\tiny$\pm$3.7e+1} & 2.5e+2 {\tiny$\pm$3.3e+3} & 1.9e+1 {\tiny$\pm$6.0e+1} & 4.8e+1 {\tiny$\pm$6.3e+2} & 2.5e+3 {\tiny$\pm$4.5e+3}  & 2.6e+3 {\tiny$\pm$4.6e+3} \\[0.15cm]
{} & {} & 1b & {} & 3.5e+0 {\tiny$\pm$1.4e+1} & 3.0e-3 {\tiny$\pm$4.5e-2} & 1.9e+1 {\tiny$\pm$6.7e+1} & 3.2e-4 {\tiny$\pm$2.9e-3} & 5.2e+0 {\tiny$\pm$9.1e+0} & 2.6e+3 {\tiny$\pm$4.6e+3} & \cellcolor{red!50} 1.3e+0 {\tiny$\pm$5.7e+0} & \cellcolor{red!50} 7.9e-6 {\tiny$\pm$1.1e-4} & 3.5e+0 {\tiny$\pm$1.4e+1} & 1.6e-5 {\tiny$\pm$1.1e-4} & 3.0e+3 {\tiny$\pm$3.5e+3} & 8.1e-2 {\tiny$\pm$1.6e+0} \\[0.15cm]
{} & {} & 2a & {} & \cellcolor{orange!50} 3.0e+0 {\tiny$\pm$4.3e+1} & \cellcolor{orange!50} 4.4e-5 {\tiny$\pm$2.5e-4} & 7.9e+0 {\tiny$\pm$1.6e+1} & 2.4e+1 {\tiny$\pm$3.8e+2} & 1.8e+1 {\tiny$\pm$2.7e+1} & 2.3e+2 {\tiny$\pm$1.6e+3} & 2.8e+1 {\tiny$\pm$5.5e+1} & 9.4e-3 {\tiny$\pm$1.7e-1} & 3.3e+0 {\tiny$\pm$6.5e+0} &  3.1e-4 {\tiny$\pm$2.8e-3} & 1.5e+1 {\tiny$\pm$2.9e+1} & 1.2e-2 {\tiny$\pm$2.7e-1} \\[0.15cm]
{} & {} & 2b & {} & 1.5e+1 {\tiny$\pm$2.7e+1} & 5.2e-2 {\tiny$\pm$5.4e-1} & 2.1e+0 {\tiny$\pm$3.5e+0} & 2.0e-1 {\tiny$\pm$2.1e+0} & 5.0e+2 {\tiny$\pm$7.7e+2} & 1.4e-3 {\tiny$\pm$7.9e-3} & 2.2e+1 {\tiny$\pm$4.4e+1} & 4.7e-2 {\tiny$\pm$3.1e-1} & 4.0e+0 {\tiny$\pm$8.7e+0} & 5.2e-3 {\tiny$\pm$3.5e-2} & 7.8e+1 {\tiny$\pm$1.1e+2} & 5.9e-3 {\tiny$\pm$3.9e-2} \\[0.1cm]
\end{tabular*}
\footnotetext{MSE results are aggregated from 5 individual runs, across 110 convection-diffusion and 100 kinematics test tasks. The DNN and PINN model configurations associated with a lower test MSE are highlighted.}
\footnotetext[1]{Initial learning rate.}
\footnotetext[2]{Prediction mode, i.e., direct DNN/PINN prediction vs. additional Tikhonov regularization (Tik-Reg) step.}
\end{sidewaystable*}

\clearpage


\begin{figure*}[h]%
\centering
\includegraphics[width=1.0\textwidth]{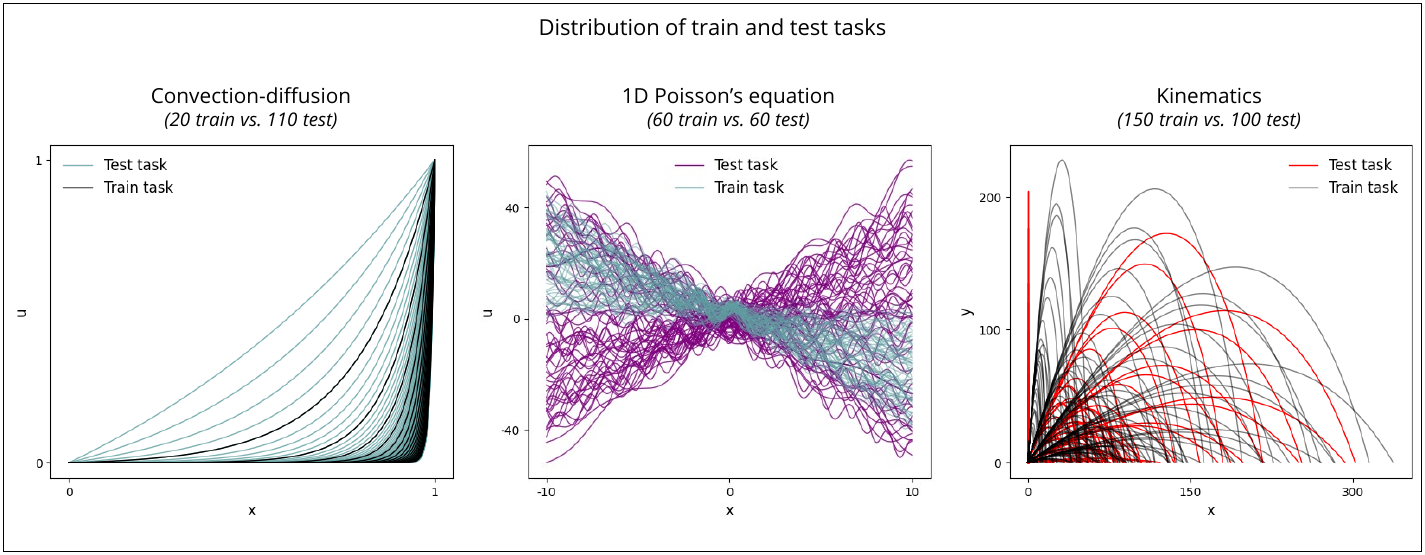}
\vspace{-0.4cm}
\caption{The distribution of convection-diffusion, 1D Poisson's, and nonlinear kinematics tasks for training and test.}\label{fig:additional_ode}
\end{figure*}


\section{Baseline SGD-Trained DNN and PINN Models Used in Ablation Study}\label{subsec:baseline}

Consider the general inputs $(x, t, \vartheta)$ for the spatial-temporal domain and task parameter $\vartheta$. The data-driven loss function of a SGD-trained DNN model computes the MSE between the DNN output $u_{\text{DNN}}(x_i, t_i, \vartheta_i)$ against the target $u_i^{label}$ over $i=1,...,n$ labelled data pooled from a batch of training tasks:
\begin{equation} 
    l_{\text{DNN}} = l_{data} = \frac{1}{n}\sum_{i=1} (u_i^{label} - u_{\text{DNN}}(x_i, t_i, \vartheta_i))^2 \label{eq:dnnloss}
\end{equation}
The loss function of a (baseline) SGD-trained PINN model is defined as:
\begin{subequations} \label{eq:pinnloss}
    \begin{align}
        l_{\text{PINN}} &= \lambda_{data} \ l_{data} + \lambda_{pde} \ l_{pde} + \lambda_{ic} \ l_{ic} + \lambda_{bc} \ l_{bc} \\
        l_{pde} &= \frac{1}{n_{pde}}\sum_{i=1} (\mathcal{N}_\vartheta[u_{\text{PINN}}(x_i^{pde}, t_i^{pde}, \vartheta_i)] - h(x_i^{pde}, t_i^{pde}))^2 \label{eq:pinnloss_pde} \\ 
        l_{ic} &= \frac{1}{n_{ic}}\sum_{i=1} (u_{\text{PINN}}(x_i^{ic}, 0, \theta_i) - u_0(x_i^{ic}))^2 \label{eq:pinnloss_ic} \\ 
        l_{bc} &= \frac{1}{n_{bc}}\sum_{i=1} (\mathcal{B}[u_{\text{PINN}}(x_i^{bc}, t_i^{bc}, \theta_i)] - g(x_i^{bc}, t_i^{bc}))^2 \label{eq:pinnloss_bc} 
    \end{align}
\end{subequations} 
such that the PINN output $u_{\text{PINN}}(x_i, t_i, \vartheta_i)$ satisfies PDE, IC, and BC for a set of training samples $(x_i^{pde}, t_i^{pde}, \vartheta_i), i=1,...,n_{pde}, (x_i^{ic}, 0, \vartheta_i), i=1,...,n_{ic}, (x_i^{bc}, t_i^{bc}, \vartheta_i), i=1,...,n_{bc}$ from the respective domain and task, in addition to minimizing the MSE from the labelled data. SGD-trained PINNs typically converge much slower than DNN because of the additional loss terms. We perform a coarse search for the loss balancing parameters $(\lambda_{data}, \lambda_{pde}, \lambda_{ic}, \lambda_{bc})$ to improve the convergence of the PINN loss.

We tested a range of learning rate schedules, e.g., an initial learning rate $= \{5\mathrm{e}{-4}, 1\mathrm{e}{-3}, 1\mathrm{e}{-2}\}$ for the first 40\% of training iterations followed by cosine decay towards $1\mathrm{e}{-6}$. Table~\ref{tab:baseline_model} and Table~\ref{tab:baseline_results} give the DNN / PINN model and training configurations and their subsequent performance on test tasks for the ablation study.

\section{Linear and Nonlinear ODE: Task Diversity} \label{subsec:additional_ablation}

Figure~\ref{fig:additional_ode} provides additional visualization results for the convection-diffusion (Problem 1), 1D Poisson's (Problem 3), and nonlinear kinematics ({Problem 6) tasks, showing the diversity of task distributions on which we have demonstrated the utility of the current study's Baldwinian-PINN.




\bibliographystyle{IEEEtran}
\bibliography{IEEEabrv,SI}